\documentclass[letterpaper]{article} % DO NOT CHANGE THIS
\usepackage{aaai2026}  % DO NOT CHANGE THIS
\usepackage{times}  % DO NOT CHANGE THIS
\usepackage{helvet}  % DO NOT CHANGE THIS
\usepackage{courier}  % DO NOT CHANGE THIS
\usepackage[hyphens]{url}  % DO NOT CHANGE THIS
\usepackage{graphicx} % DO NOT CHANGE THIS
\usepackage{natbib}  % DO NOT CHANGE THIS AND DO NOT ADD ANY OPTIONS TO IT
\usepackage{caption} % DO NOT CHANGE THIS AND DO NOT ADD ANY OPTIONS TO IT
\usepackage{algorithm}
\usepackage{algorithmic}

\usepackage{newfloat}
\usepackage{listings}
\DeclareCaptionStyle{ruled}{labelfont=normalfont,labelsep=colon,strut=off} % DO NOT CHANGE THIS
\floatstyle{ruled}
\newfloat{listing}{tb}{lst}{}
\floatname{listing}{Listing}
\usepackage{url}            % simple URL typesetting
\usepackage{amsfonts}       % blackboard math symbols
\usepackage{nicefrac}       % compact symbols for 1/2, etc.
\usepackage{microtype}      % microtypography
\usepackage{xcolor}         % colors
\usepackage{subfigure}
\usepackage{booktabs} % for professional tables
\usepackage{enumitem}
\usepackage{multirow}

\usepackage{svg}
\usepackage{adjustbox}
\usepackage{amsmath}
\usepackage{subcaption}

\title{Near-optimal Linear Predictive Clustering in Non-separable Spaces via MIP and QPBO Reductions}

\author {
    Jiazhou Liang\equalcontrib\textsuperscript{\rm 1},
    Hassan Khurram\equalcontrib\textsuperscript{\rm 1},
    Scott Sanner\textsuperscript{\rm 1 2}
}
\affiliations {
    \textsuperscript{\rm 1} University of Toronto\\
    \textsuperscript{\rm 2} Vector Institute for Artificial Intelligence\\
    joe.liang@mail.utoronto.ca, hassan.khurram@mail.utoronto.ca, ssanner@mie.utoronto.ca
}

\begin{document}

\maketitle

\begin{abstract}
Linear Predictive Clustering (LPC) partitions samples based on shared linear relationships between feature and target variables, with numerous applications including marketing, medicine, and education. Greedy optimization methods, commonly used for LPC, alternate between clustering and linear regression but lack global optimality. While effective for separable clusters, they struggle in \emph{non-separable} settings where clusters overlap in feature space.  In an alternative constrained optimization paradigm, previous work formulated LPC as a Mixed-Integer Program (MIP), ensuring global optimality regardless of separability but suffering from poor scalability.  This work builds on the constrained optimization paradigm to introduce two novel approaches that improve the efficiency of global optimization for LPC. By leveraging key theoretical properties of separability, we derive near-optimal approximations with provable error bounds, significantly reducing the MIP formulation’s complexity and improving scalability. Additionally, we can further approximate LPC as a Quadratic Pseudo-Boolean Optimization (QPBO) problem, achieving substantial computational improvements in some settings. Comparative analyses on synthetic and real-world datasets demonstrate that our methods consistently achieve near-optimal solutions with substantially lower regression errors than greedy optimization while exhibiting superior scalability over existing MIP formulations.
\end{abstract}

\begin{links}
    \link{Code}{github.com/D3Mlab/LPC-NS}
     \link{Extended version}{arxiv.org/abs/2511.10809}
\end{links}

\section{Introduction}
Linear Predictive Clustering (LPC) clusters samples into  $K$ distinct groups, each characterized by a unique set of linear coefficients shared among the samples within the group, thereby capturing clusterwise patterns and variations in linear relationships.
The versatility of LPC enables its application across diverse fields. In marketing, LPC facilitates market segmentation by predicting customer purchase behavior~\cite{WEDEL198945}. In medicine, LPC supports patient stratification for predictive health analytics, enhancing personalized treatment strategies~\cite{Ntani2020Consequences}. Similarly, in education, LPC contributes to personalized learning by analyzing student performance and tailoring educational approaches based on outcomes predicted through linear models~\cite{Naik2017Predicting}.

\begin{figure}
\centering
\includegraphics[width=\linewidth]{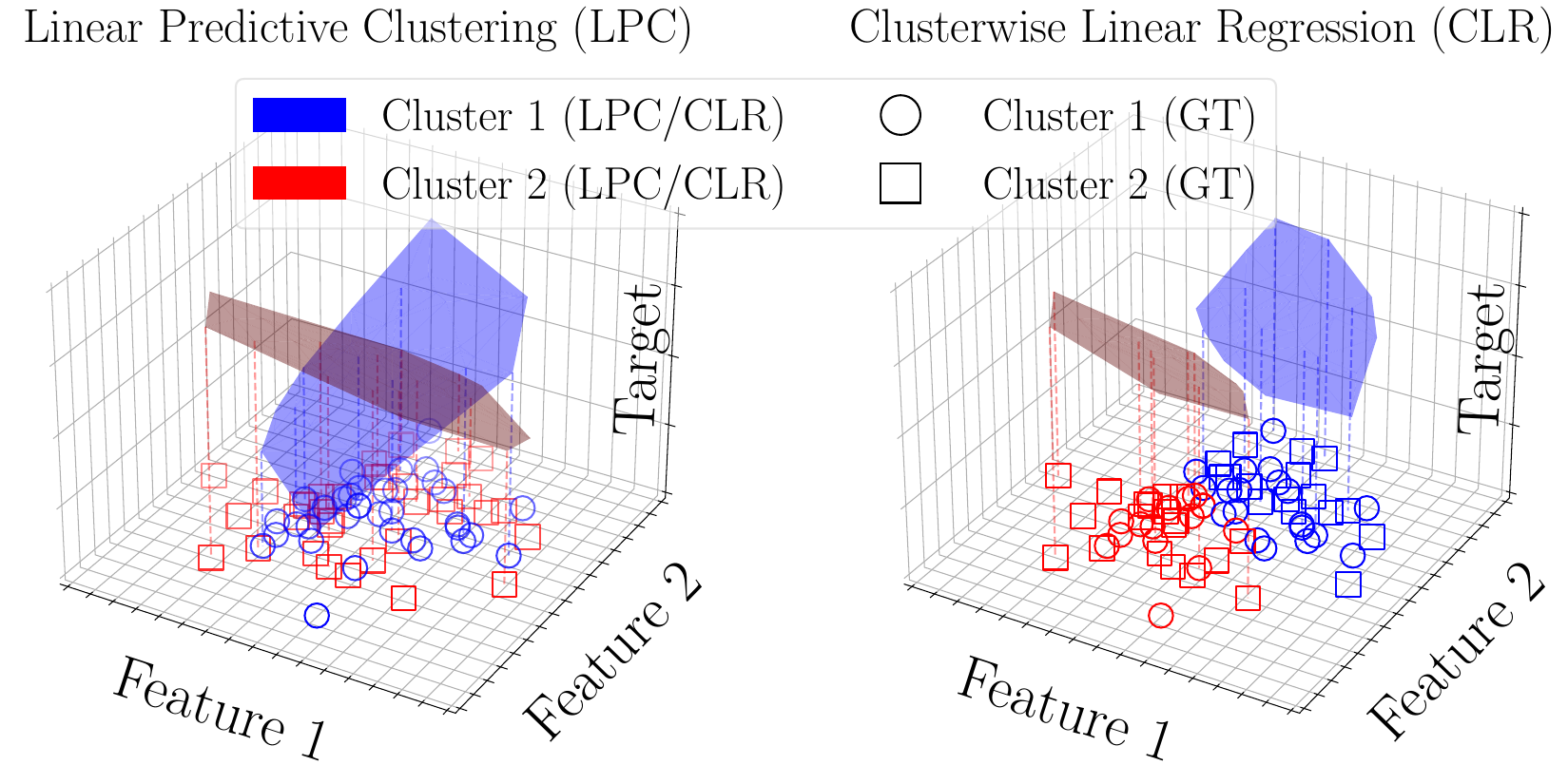}
\caption{ The plot illustrates a case where the feature variables (horizontal axes) from samples belonging to two different ground truth clusters (circles and squares) overlap, making them non-separable in the feature space. However, each cluster follows a distinct linear relationship (hyperplane) with the target variable (vertical axis). This non-separability renders CLR (cf. \textit{right})—which clusters solely based on feature variables—ineffective: it assigns clusters (shown in blue and red) that mix the ground truth labels. In contrast, LPC (cf. \textit{left}) recovers the ground truth linear predictor assignment.
}
\label{fig:illustration_of_clr}
\end{figure}

In a special case of the LPC problem, feature variables may be \textit{non-separable}, meaning that samples from different clusters overlap in the feature space. Specifically, when the means and uncentered covariance of feature variables are highly similar across clusters, clustering based solely on feature variables becomes ineffective. As shown in Fig.~\ref{fig:illustration_of_clr} (left), although each cluster follows a distinct relationship with the target variable (hyperplanes on the vertical axis), samples from the two ground truth clusters (shown as circles and squares) exhibit significant overlap in their feature space (horizontal axes). This non-separability renders simple \emph{clustering-then-regression} approaches
highly impractical. As illustrated in  Fig.~\ref{fig:illustration_of_clr} (right), applying an external clustering method to feature variables in Clusterwise Linear Regression (CLR) produces two partitions (blue and red) as a mixture of samples from two ground truth clusters (circles and squares), which fail to recover the ground truth linear predictor assignment. This further demonstrates the limitations of traditional clustering-based approaches in the non-separable setting.

Existing LPC methods for \emph{non-separable feature spaces} fall into two categories. The first, Iterative (Greedy) Optimization~\citep{spath1979algorithmus, manwani2015k, wang2020interpretable}, alternates between updating regression coefficients and cluster assignments. While this approach has been widely adopted due to its scalability, it is highly sensitive to initialization, lacks global optimality, and is prone to convergence at suboptimal local minima~\citep{chembu2023generalized}. 
The second, Constrained Optimization~\citep{bertsimas2007classification, carbonneau2012extensions, zhu2012clusterwise, chembu2023generalized},  formulates LPC as a Mixed-Integer Programming (MIP) problem, jointly optimizing cluster assignments and regression coefficients. This guarantees global optimality by minimizing total regression loss but suffers from significant scalability limitations due to its computational complexity.

\paragraph{Proposed Contribution.}  To the best of our knowledge, developing an approach that balances the scalability of greedy optimization with the global optimality of MIP-based formulations for LPC in non-separable feature spaces remains an open challenge.  Hence, to address these challenges, we make the following key contributions in this work:

\begin{itemize}
\item We build on the global optimization MIP approach to LPC and propose a novel near-optimal approximation of regression coefficients by leveraging fundamental properties of non-separability. This insight allows us to substitute and ablate the regression component from the MIP LPC constrained optimization, thereby significantly reducing the size of the MIP formulation. This leads to our first contribution of LPC-NS-MIP for efficient LPC in non-separable (NS) feature spaces.

\item We derive error bounds on LPC-NS-MIP that prove the optimality of our approach under non-separability conditions on the uncentered covariances of each cluster.

\item We further transform LPC-NS-MIP into a Quadratic Pseudo-Boolean Optimization (QPBO) formulation.  Empirically, LPC-NS-QPBO offers substantial scalability improvements while maintaining near-optimal performance in the setting of two clusters.

\item We conduct rigorous experiments demonstrating that LPC-NS-MIP and LPC-NS-QPBO consistently outperform the greedy optimization approach, achieving significantly better LPC objective optimization across datasets with varying noise, dimensionality, and outliers, while scaling more efficiently than existing globally optimal MIP methods.
\end{itemize}

\section{Related Work}
\paragraph{LPC under Constrained Optimization} was first introduced by~\citet{bertsimas2007classification} as CRIO, which formulates the problem as a MIP problem to identify optimal cluster assignments that minimize the within-cluster regression Sum of Absolute Errors (SAE). This formulation guarantees deterministic cluster assignments with global optimality.~\citet{zhu2012clusterwise} introduces a more outlier-robust variation, and~\citet{carbonneau2012extensions} extends CRIO with a Sum of Squared Errors (SSE) objectives.  We extend and compare to these global optimization methods.

\paragraph{LPC under Greedy Optimization} was first proposed by~\citet{spath1979algorithmus} and later extended by~\citet{chembu2023generalized}, drawing inspiration from the Majorization-Minimization (MM) algorithm~\citep{hunter2004tutorial}.
These methods employ an iterative approach, where cluster assignments are fixed to compute cluster-specific regression lines, followed by updating cluster assignments to minimize the clusterwise regression loss. Cluster updates are performed either by pairwise sample exchanges~\citep{spath1979algorithmus} or by re-assigning all samples~\citep{chembu2023generalized}, repeating the process until convergence.  Greedy methods are fast, but prone to local optima as we show empirically.

\paragraph{Clusterwise Linear Regression (CLR)} CLR~\citep{BradleykPlane, hota2009simple, manwani2015k, feng2016spatial, corizzo2019dencast} fundamentally differs from LPC in its approach to cluster assignments. While LPC assigns data to clusters arbitrarily, focusing solely on the objective of linear regression, CLR often uses external clustering definitions informed by prior knowledge.
For instance, K-plane regression ~\citep{manwani2015k, da2017combining, wang2020interpretable} extends the $K$-means algorithm~\citep{lloyd1982least} into a regression context by leveraging Mean Square Sum of Clusters (MSSC) error.

This approach struggles with non-separable feature spaces as feature variables cannot form distinct clusters.
As shown in Fig.~\ref{fig:illustration_of_clr} (right), the distribution of cluster feature variables may conflict with MSSC objectives that require some degree of separability.  Since CLR cannot address this non-separable case, it does not address the main focus of our work.

\paragraph{Latent Class Regression (LCR)}  
Unlike LPC, which aims to identify deterministic cluster assignments with minimum regression loss, Latent Class Regression (LCR)~\citep{wedel1994review} can be viewed as a generalization of CLR that employs an explicit discrete Finite Mixture Model (FMM) for each latent class~\citep{quandt1972new, desarbo1988maximum, cosslett1985serial, hamilton1990analysis}. 
However, in non-separable settings, where multiple latent classes share the same modality, LCR fails to distinguish clusters effectively for the same reasons that CLR failed in Fig.~\ref{fig:illustration_of_clr} (right). Since LCR cannot handle the non-separable case, it does not address the main focus of our work.

\section{Methodology}
\subsection{Preliminary Definitions and MIP formulation with Global Optimality}
We first define the objective of LPC and introduce the state-of-the-art MIP formulations~\citep{bertsimas2007classification}.  

Let \( \mathbf{X} \in \mathbb{R}^{N \times (D + 1)} \) denote \( D \) features variables with bias term for a dataset with \( N \) samples, and let \( \mathbf{y} \in \mathbb{R}^{N \times 1} \) represent the corresponding target feature. Cluster assignments are encoded by \( \mathbf{Z}_k \in \mathbb{Z}^{N \times N} \), a binary diagonal matrix where \( \mathbf{Z}_{k,i,i} = 1 \) if sample \( i \) is assigned to cluster \( k \) and 0 otherwise. The regression coefficients and bias for cluster \( k \) are given by \( \mathbf{w}_k \in \mathbb{R}^{D + 1} \).  

The Regularized Linear Least Squares (RLS) 
estimates a single set of coefficients \( \mathbf{w} \) minimizing the SSE between \( \mathbf{y} \) and its prediction \( \mathbf{Xw} \)  with \(\ell_2\) regularization (a.k.a. Ridge Regression).
LPC extends from this objective by partitioning \( \mathbf{X} \) into \( K \) disjoint clusters, where \( K \) is predefined based on prior knowledge. The objective function of LPC is formulated as:  
\begin{equation} \label{eq:lpc_objective}
    \min_{\mathbf{Z}, \mathbf{w}} \sum_{k=1}^K \left( \| \mathbf{Z}_k \mathbf{y} - \mathbf{Z}_k \mathbf{X} \mathbf{w}_k \|_2^2 + \lambda \| \mathbf{w}_k \|_2^2 \right).
\end{equation}  
Since \( \mathbf{Z}_{k,i,i} = 1 \) only if sample \( i \) is assigned to cluster \( k \), the residual \( \mathbf{y}_i - \mathbf{X}_i \mathbf{w}_k \) is nonzero only for assigned samples. Consequently, the SSE for \( \mathbf{w}_k \) is computed exclusively over samples in cluster \( k \).  LPC aims to determine the optimal \( \mathbf{Z}_k \) and \( \mathbf{w}_k \) such that the total SSE across clusters is minimized.

Equation~\eqref{eq:lpc_objective} can be solved using MIP with optimization variables $\mathbf{Z}$ and $\mathbf{w}$~\citep{bertsimas2007classification,zhu2012clusterwise,carbonneau2012extensions}, ensuring a globally optimized objective. The MIP formulation is expressed as:
\begin{align} \label{eq:clr_miqp_sse}
    \min_{\mathbf{Z}, \mathbf{w}} & \sum_{k=1}^K \left(\sum_{i=1}^n \delta_{i,k}^2 + \lambda \| \mathbf{w}_k \|_2^2 \right) \\
    \text{s.t.  } &  \delta_{i} \leq (\mathbf{y}_i - \mathbf{X}_i\mathbf{w}_k) + M (1 - \mathbf{Z}_{k,i,i}) \nonumber \\ & \qquad \forall i \in \{1,\dots,N\} ; k \in \{1,\dots,K\}  \label{eq:big-m-1} \\
    & \delta_{i} \geq (\mathbf{y}_i - \mathbf{X}_i\mathbf{w}_k) - M (1 - \mathbf{Z}_{k,i,i}) \nonumber \\ & \qquad \forall i \in \{1,\dots,N\} ; k \in \{1,\dots,K\} \label{eq:big-m-2} \\ 
    & \textstyle \sum_{k=1}^K \mathbf{Z}_{k,i,i} = 1, \quad \forall i \in \{1,\dots,N\} \label{eq:allone} \\
    &\delta_{i} \in \mathbb{R}; \quad \mathbf{Z}_{k,i,i} \in \{0,1\};  \quad  W_{jk} \in \mathbb{R}  \nonumber \\
    &  \quad  \forall i \in \{1,\dots,N\} j \in \{1,\dots,D + 1\} ; k \in \{1,\dots,K \nonumber \} 
\end{align}
Here, \( \delta \) is a continuous auxiliary variable used to linearize the clusterwise regression error, i.e., \(  \mathbf{Z}_{k,i,i} = 1 \rightarrow \delta_{i} = (\mathbf{y}_i - \mathbf{X}_i\mathbf{w}_k) \), via the Big-M constraint~\citep{dantzig2016linear}, where \( M \) is a large positive scalar. Constraints~\eqref{eq:big-m-1} and~\eqref{eq:big-m-2} ensure that the residual of sample \( i \) with respect to \( \mathbf{w}_k \) contributes to the objective only when \( i \) is assigned to cluster \( k \). Constraint~\eqref{eq:allone} enforces that each sample \( i \) is assigned to exactly one cluster, ensuring that the \( K \) clusters are mutually exclusive and exhaustively cover all samples.

\subsection{Optimization of Regression Coefficients $\mathbf{w}$}

We observed that jointly optimizing $\mathbf{Z}$ and $\mathbf{w}$ in \eqref{eq:clr_miqp_sse} significantly limits the scalability of LPC. To improve efficiency, we aim to reduce the number of optimization variables in this formulation.
 
Given a known \( \mathbf{Z}_k \), the corresponding regression coefficients \( \mathbf{w} \) can be optimized implicitly via the closed-form solution as follows:
\begin{equation}\label{eq:weight}
    \mathbf{w}_k = \underbrace{\left( \mathbf{X}^{\top} \mathbf{Z}_k \mathbf{X} + \lambda  \mathbf{I} \right)^{-1}}_{\mathbf{A}_k} \mathbf{X}^{\top} \mathbf{Z}_k \mathbf{y},
\end{equation}
Substituting  \eqref{eq:weight} into  \eqref{eq:lpc_objective}, the optimal \( \mathbf{w} \) can be directly computed using this implicit formulation, eliminating the need for two sets of optimization variables in the MIP formulation. 

However, the matrix inversion in $\mathbf{A}_k$ depends on the optimization variable \( \mathbf{Z}_k \), which is infeasible to formulate in an MIP problem. The term \( \mathbf{X}^{\top} \mathbf{Z}_k \mathbf{X} \) in \( \mathbf{A}_k \) represents the clusterwise uncentered covariance of the feature variables \( \mathbf{X} \), computed using only the samples assigned to cluster \( k \). 

The definition of \emph{non-separable} \( \mathbf{X} \) suggests that the uncentered covariance structure $\mathbf{X}^{\top}\mathbf{X}$ across different subsets of samples often remains similar, cf. Fig.~\ref{fig:illustration_of_clr}(left), where the non-separable setting is precisely the case where the clusters have \emph{matching} uncentered covariance.  
Consequently, while computing the matrix inverse 
for each cluster in \( \mathbf{A}_k \) is infeasible to formulate as a MIP problem, fortuitously, it may actually be unnecessary!

To address this, we propose an approximation for \( \mathbf{A}_k \), denoted as \( \mathbf{A}^\ast \), defined as:  
\begin{equation}\label{Wast}
    \mathbf{w}_k^* = \underbrace{\left( \alpha_k\mathbf{X}^{\top}\mathbf{X} + \lambda  \mathbf{I} \right)^{-1}}_{\mathbf{A^*}} \mathbf{X}^{\top} \mathbf{Z}_k \mathbf{y},
\end{equation}

\noindent where $\alpha_k$ is a hyperparameter reflecting the relative size of cluster $k$. Under balanced cluster sizes, this reduces to $\alpha_k = 1/K$. For imbalanced clusters, a multi-hop tuning procedure yields a more suitable value (cf. App.D for details).

Eliminating \( \mathbf{Z}_k \) from \( \mathbf{A}_k \) enables precomputing \( \mathbf{A}^\ast \), reducing the optimization of \( \mathbf{w} \) to a matrix multiplication, leading to a simplified MIP formulation.  Specifically, substituting the  \( \mathbf{w}_k^\ast \)  back into \eqref{eq:lpc_objective}, we derive a quadratic form solely in terms of the cluster assignment variables \( \mathbf{Z}_k \):
\begin{equation} \label{eq:lpc_objective_z}
    \min_{\mathbf{Z}} \sum_{k=1}^K \left( \| \mathbf{Z}_k \mathbf{y} - \mathbf{Z}_k \mathbf{X} \mathbf{A}^\ast \mathbf{X}^{\top} \mathbf{Z}_k \mathbf{y} \|_2^2 + \lambda \| \mathbf{A}^\ast \mathbf{X}^{\top} \mathbf{Z}_k \mathbf{y} \|_2^2 \right)
\end{equation}
Here, the only variable requiring optimization is the binary indicator \( \mathbf{Z}_k \), while the regression coefficients \( \mathbf{w} \) are implicitly optimized as given in \eqref{Wast}.  
This transformation successfully reduces the MIQP formulation of LPC into a higher-order Pseudo-Boolean Optimization (PBO) problem, which we reduce to a Quadratic PBO (QPBO) form in Sec.~\ref{sec:qpbo}. 

\emph{With this critical insight, we have shown that this leads to a substantial reduction in the complexity of the MIP encoding due to the ability to directly solve for $w_k^\ast$ and ablate these optimization variables without sacrificing optimality under the separability assumption.  However, before we proceed further under this assumption, we first investigate settings where this approximation is sensible.}

\subsection{Approximation Error of $\mathbf{w}^\ast$ }
\label{sec:error}

Although the non-separable nature of \( \mathbf{X} \) provides insight into the approximation of \( \mathbf{A}^\ast \), it is crucial to both theoretically and empirically assess the bound of the approximation error in the LPC objective introduced by approximating \( \mathbf{w} \) as \( \mathbf{w}^\ast \) in \eqref{Wast}.

Let \( \mathcal{E}_{\text{obj}} \) denote the sum of differences between the globally optimal linear regression error objective in cluster \( k \) (\( O_k \)) and the linear regression error objective using \( \mathbf{w}^\ast \) from  \eqref{Wast} (\( O_k^\ast \)), i.e.,  
\(
\mathcal{E}_{\text{obj}} = \sum_{k=1}^K \left( O_k^\ast - O_k \right).
\) 
This term quantifies the approximation error introduced in the LPC objective by using \( \mathbf{w}^\ast \) instead of the globally optimal solution. Assuming \( \lambda > 0 \), the upper bound of \( \mathcal{E}_{\text{obj}} \) is as follows (cf. App. A for the detailed error derivation):
\begin{equation}
    \label{upper_bound_approximation_error}
    \mathcal{E}_{\text{obj}} \leq \sum_{k=1}^{K} \left( \frac{3\|\mathbf{y}_k\|^2_2 \| \mathbf{X}_k\|^2_2 \|\alpha_k\mathbf{X}^\top\mathbf{X} - \mathbf{X}^\top\mathbf{Z}_k\mathbf{X}\|_2}{2\lambda^2} \right),
\end{equation}

\begin{table*}
\centering
\small
\renewcommand{\arraystretch}{1.5} % Adjust row height for better readability
\setlength{\tabcolsep}{6pt} % Adjust column separation
\begin{tabular}{@{}lcccc@{}}
\toprule
\textbf{Types} & \textbf{Identical} & \textbf{Non-Separable} & \textbf{Semi-Separable} & \textbf{Separable} \\ 
\midrule
\textbf{Samples} & 
\adjustbox{valign=m}{\includegraphics[width=0.20\textwidth]{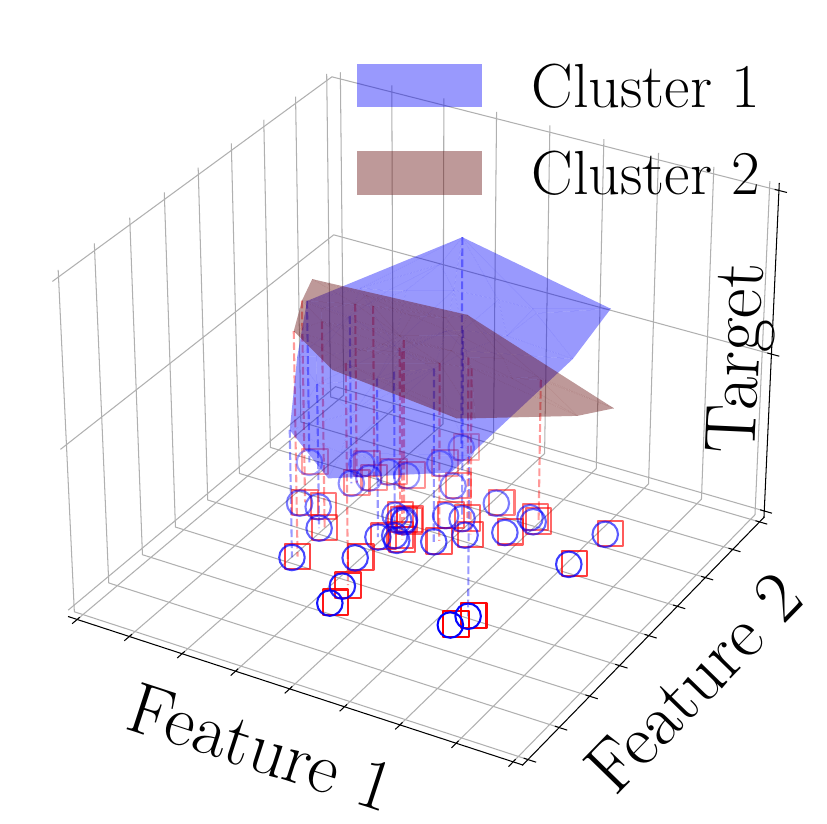}} & 
\adjustbox{valign=m}{\includegraphics[width=0.20\textwidth]{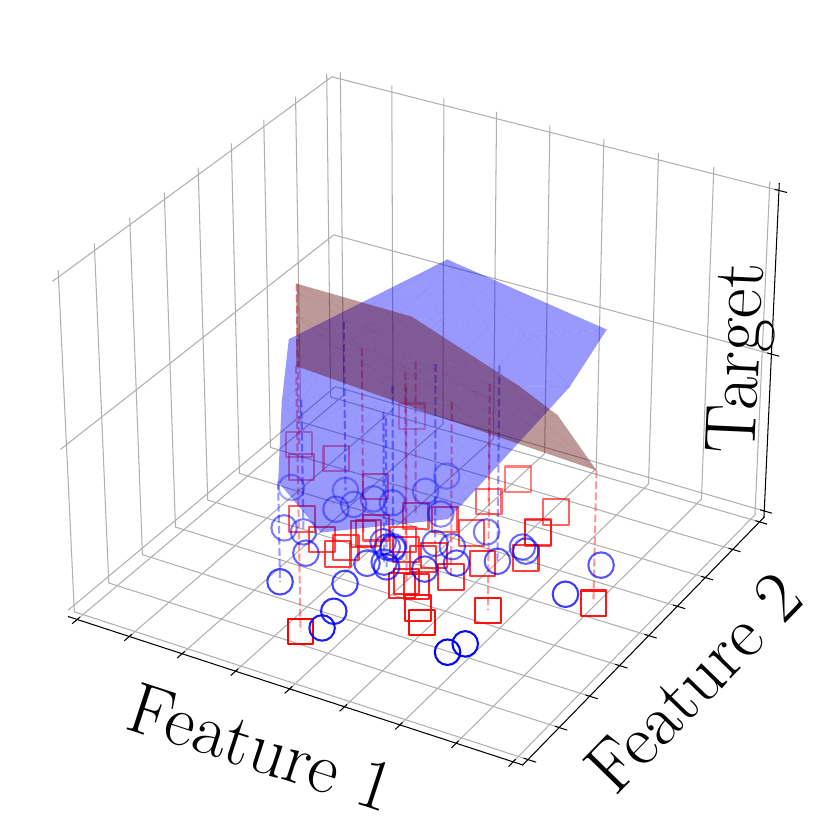}} & 
\adjustbox{valign=m}{\includegraphics[width=0.20\textwidth]{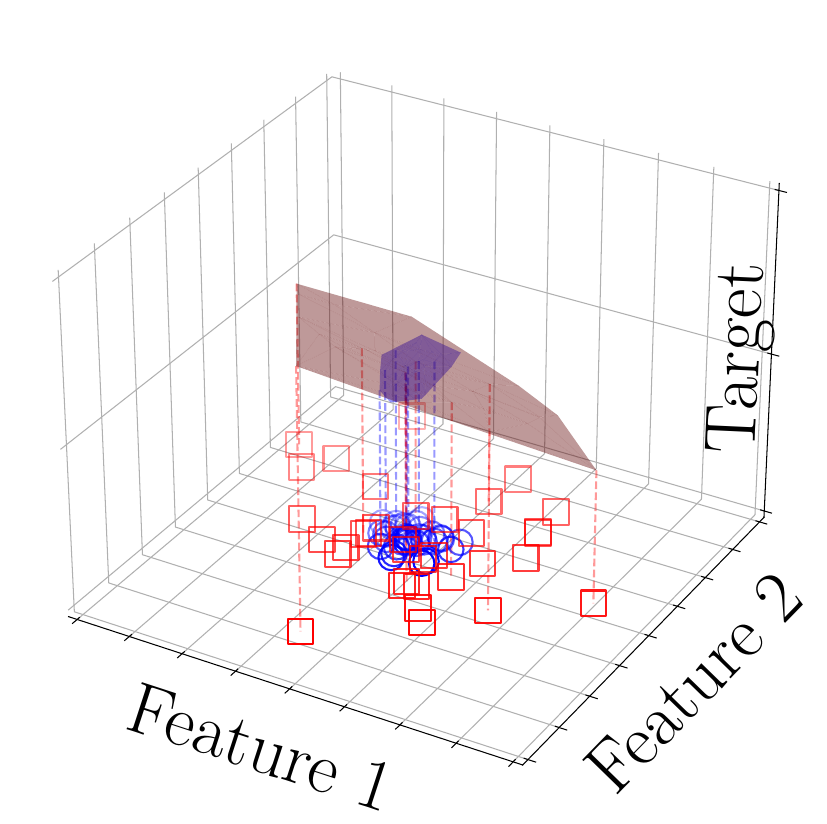}} & 
\adjustbox{valign=m}{\includegraphics[width=0.20\textwidth]{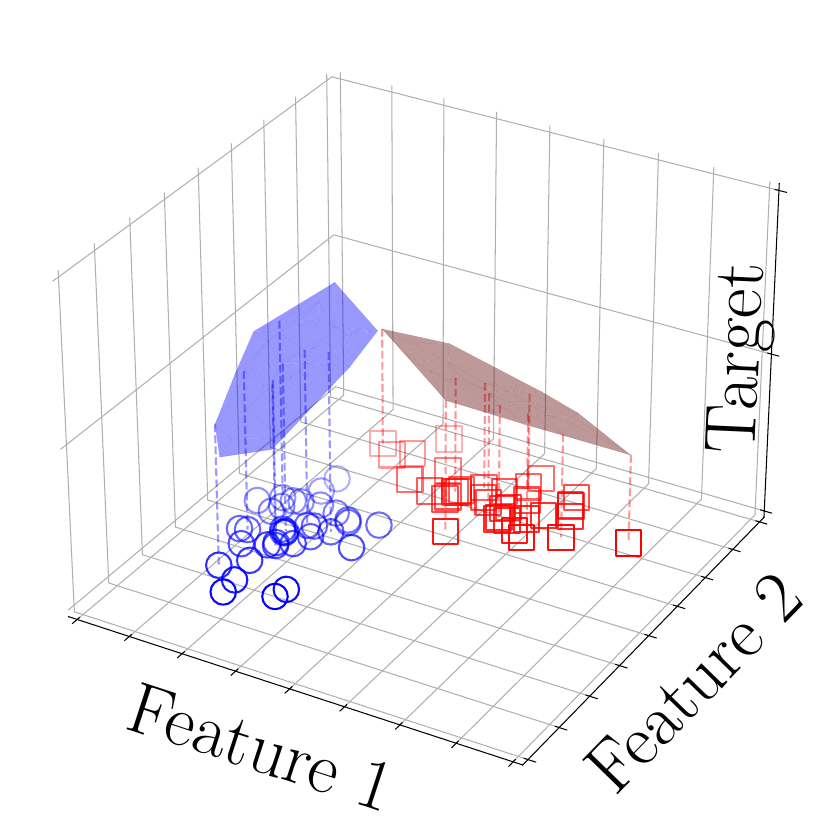}} \\ 
\midrule
{\Large$\mathcal{E}_{\text{sep}}$} & 0.0 & 60.4043 & 193.791 & 666.579 \\ 
{\Large $\mathcal{E}_{\text{obj}}^{\mathbf{w}^\ast}$} & $0.01 \pm 0.01$ & $0.0208 \pm 0.0054$ & $2.7080 \pm 0.4973$ & $2.6250 \pm 0.4868$ \\ 
{\Large$\mathcal{E}_{\text{obj}}^{\text{CLR}}$}  & $ 18.4862 \pm 0.4441$ & $72.2481 \pm 0.8360$ & $9.4433 \pm 0.3633$ & $0.5996 \pm 0.0316$ \\
\bottomrule
\end{tabular}
\caption{Samples from two clusters exhibits two distinct linear relationships ($\mathbf{w}_1 = [2,4,-5]$ and $\mathbf{w}_2 =[-2,-4,5]$) under four types of feature variables in 2-dimensional space (\(\mathbf{X} \in \mathbb{R}^{N \times 2}\)), ordered by ascending values of  $\mathcal{E}_{\text{sep}}=\sum_{k=1}^2\left\|\alpha_k\mathbf{X}^\top\mathbf{X} - \mathbf{X}^\top\mathbf{Z}_k\mathbf{X}\right\|_2$.  
The difference between the globally optimal objective (\(O\)) and optimal objective via $\mathbf{w}^\ast$ approximation (\( O^{\mathbf{w}^\ast} \)) is computed as  
\( \mathcal{E}_{\text{obj}}^{\mathbf{w}^\ast} = O^{\mathbf{w}^\ast} - O \) with 95\% confidence interval. The clusterwise regression (CLR) objective (\( O^{\text{CLR}} \)) and its deviation from \( O \), denoted as  
\( \mathcal{E}_{\text{obj}}^{\text{CLR}} \), is obtained by K-Means clustering on \(\mathbf{X}\) and Ridge regression to each cluster.}
\label{tab:four_cases}
\end{table*}

In \eqref{upper_bound_approximation_error}, the terms \( \|\mathbf{y}_k\|^2_2 \) and \( \| \mathbf{X}_k\|^2_2 \) correspond to the norms of the target and feature variables of samples within each cluster. Except for degenerate cases of purely null data, these values cannot be zero for non-empty clusters. 

The term \( \left\|\alpha_k\mathbf{X}^\top\mathbf{X} - \mathbf{X}^\top\mathbf{Z}_k\mathbf{X}\right\|_2 \) in \eqref{upper_bound_approximation_error} quantifies the difference in uncentered covariance between all samples (\( \mathbf{X}^\top\mathbf{X} \)) and those assigned to cluster \( k \) (i.e., \( \mathbf{X}^\top\mathbf{Z}_k\mathbf{X} \)). 

When the feature variables are perfectly non-separable (i.e., identical across clusters) and cluster sizes are balanced (\( N/K \) samples per cluster with $\alpha_k = 1/K$) , as illustrated in the first plot of Tab.~\ref{tab:four_cases},  
\(
\left\|\alpha_k\mathbf{X}^\top\mathbf{X} - \mathbf{X}^\top\mathbf{Z}_k\mathbf{X}\right\|_2 = 0, \quad \forall k \in \{1, \dots, K\}.
\)
Thus, the approximation error satisfies \( \mathcal{E}_{\text{obj}} \leq 0 \), ensuring perfect recovery of the globally optimal objective. For imbalanced cluster settings, $\alpha_k$ can be iteratively tuned to adapt to varying cluster sizes (cf. App.~D for details). 

Note that this 0-bound assumes that the $Z_k$ obtained from optimizing \eqref{eq:lpc_objective_z} is aligned with the ground truth cluster assignment. Nevertheless, \eqref{eq:lpc_objective_z} is inherently robust to suboptimal $Z_k$ since it produces poor regression coefficients with high cluster-wise SSE, which are penalized in the objective.

Tab.~\ref{tab:four_cases} empirically evaluates the deviation from the globally optimal objective when using $\mathbf{w}^\ast$, denoted as \( \mathcal{E}_{\text{obj}}^{\mathbf{w}^\ast} \), across four synthetic datasets with \( K = 2, N = 50 \) (cf. Sec.~\ref{sec:experiment_setup}). Each dataset maintains fixed parameters and orthogonal regression coefficients $\mathbf{w}_k$ between clusters while varying \( \mathbf{X} \), representing different levels of separability:  
\begin{equation}
\mathcal{E}_{\text{sep}} = \sum_{k=1}^K \left\|\alpha_k\mathbf{X}^\top\mathbf{X} - \mathbf{X}^\top\mathbf{Z}_k\mathbf{X}\right\|_2 \, .
\end{equation}  
When \( \mathbf{X} \) is perfectly non-separable (\(\mathcal{E}_{\text{sep}} = 0\)), the approximation error \( \mathcal{E}_{\text{obj}}^{\mathbf{w}^\ast} \) is zero. More generally, in non-separable cases (second column of Tab.~\ref{tab:four_cases}), \( \mathbf{w}^\ast \) remains a near-optimal approximation of \( \mathbf{w} \).

Two additional cases on the right of Tab.~\ref{tab:four_cases} illustrate: \emph{semi-separable}, where two clusters overlap in $X$ but differ in uncentered covariances, and \emph{separable}, where clusters are distinct in $X$ and have different covariance. In both cases, as feature variables become more separable, the approximation error of $\mathbf{w}^\ast$ increases, resulting in a larger optimality gap. However, these separable cases already have well-established solutions, such as CLR, achieving near-optimal solutions in the separable setting; thus, we remark that we specifically focus on the more challenging \emph{non-separable} setting. 

\subsection{QPBO formulation of Cluster Assignment $\mathbf{Z}$}
\label{sec:qpbo}

Although \eqref{eq:lpc_objective_z} can technically be rewritten as a cubic PBO problem (cf. App.~B), no modern constrained optimization solvers currently provide scalable solutions due to the cubic growth of optimization variables \( \mathbf{Z} \). Consequently, in practice, a MIP formulation incorporating a continuous auxiliary variable \( \boldsymbol{\delta} \) (cf. \eqref{eq:clr_miqp_sse}) offers superior scalability compared to direct PBO formulations.

However, LPC-NS-MIP can be approximated as a Quadratic Pseudo-Boolean Optimization (QPBO) problem, termed LPC-NS-QPBO. We first reformulate and simplify \eqref{eq:lpc_objective_z} as a maximization problem after expanding the square (cf. App.~B for a detailed derivation):  
\begin{equation}  \label{simplied_objective}
        \max_{\mathbf{Z}} \sum_{k=1}^K \! \left[ \mathbf{y}^\top \mathbf{Z}_k \mathbf{X}^{\top}  A^\ast [2\mathbf{I} \! - \! \underbrace{(\mathbf{X}^{\top} \mathbf{Z}_k \mathbf{X} + \lambda  \mathbf{I})}_{\mathbf{A}_k^{-1}}A^\ast]\mathbf{X}^{\top}  \mathbf{Z}_k\mathbf{y} \right]
\end{equation}
From \eqref{eq:weight}, we recognize the form of \(\mathbf{A}_k^{-1}\) and then approximate it as \( A^{\ast-1} \) under the non-separable assumption using \eqref{Wast}. 
Then we simplify \mbox{$[2\mathbf{I} - A^{\ast-1}A^\ast] = [2\mathbf{I}-\mathbf{I}]=\mathbf{I}$}, yielding the final reduced formulation for LPC-NS-QPBO:  
\begin{align} \label{eq:lpc-ns-qpbo}
    \max_{\mathbf{Z}} & \textstyle \sum_{k=1}^K \left[ \mathbf{y}^\top \mathbf{Z}_k \mathbf{X}^{\top}  A^\ast \mathbf{X}^{\top}  \mathbf{Z}_k\mathbf{y} \right] \\
    \text{s.t.}
    & \textstyle \sum_{k=1}^K \mathbf{Z}_{k,i,i} = 1,  \mathbf{Z}_{k,i,i} \in \{0,1\}; \nonumber \\ &  \quad \forall i \in \{1,\dots,N\};  k \in \{1,\dots,K\}
\end{align}

LPC-NS-QPBO effectively reduces LPC-NS-MIP into an approximate QPBO formulation with binary optimization variable \( \mathbf{Z} \). Moreover, since \( \mathbf{X}^{\top}  A^\ast \mathbf{X}^{\top} \) does not involve any optimization variables, it can be precomputed, further simplifying the optimization process.

\subsection{Refitting Regression Coefficients  $\mathbf{w}$ using Optimal Clustering Assignment $\mathbf{Z}$  }

After obtaining the optimal \( \mathbf{Z} \) using LPC-NS-QPBO or LPC-NS-MIP, optimal \( \mathbf{w}_k \) can be recovered using \eqref{eq:weight}. Thus, this refitting process, denoted as \( \mathbf{w}_k^{\text{refit}} \) for all \( k \in \{1, \dots, K\} \), is given by:  
\begin{equation}\label{refit}
    \mathbf{w}_k^{\text{refit}} = \left( \mathbf{X}^{\top} \mathbf{Z}_k \mathbf{X} + \lambda \mathbf{I} \right)^{-1} \mathbf{X}^{\top} \mathbf{Z}_k \mathbf{y}.
\end{equation}
This step ensures that the refitted regression coefficients \( \mathbf{w}_k^{\text{refit}} \) are optimally aligned with the final cluster label assignments. 

\section{Experiments}
\subsection{Research Questions}
To evaluate the effectiveness of our proposed methods, LPC-NS-MIP and LPC-NS-QPBO, against existing approaches -- Greedy Optimization (Greedy) and globally optimal methods (GlobalOpt) -- we conduct experiments focusing on three key aspects: (a) deviation of each method from the GlobalOpt LPC objective, (b) recovery performance of regression coefficients $\mathbf{w}^\text{gt}$ and clustering assignments $\mathbf{Z}^\text{gt}$ in the synthetic datasets with known ground truth, and (c) optimization times vs. data sample size. The research questions (RQs) guiding our evaluation are as follows:
\begin{itemize}
\item \textbf{RQ1: Time vs. Error Scalability Trade-off Analysis.}
How do objective error and optimization times change as the number of samples increases? 
\item \textbf{RQ2: Performance Analysis and Algorithm Properties on Synthetic Datasets.}  
For synthetic datasets with known ground truth, how do different methods compare in terms of objective and accuracy? We evaluate these metrics for different $K$ while varying feature dimensionality, noise levels, and proportion of target variable outliers.
\item \textbf{RQ3: Performance Analysis on Real Datasets.}  
How do different methods compare in terms of objective error on real datasets, specifically curated to contain two distinct clusters with differing regression coefficients? 
\end{itemize}

\subsection{Experimental Setup} \label{sec:experiment_setup}

\paragraph{Synthetic Datasets (SD)}  
To simulate a non-separable observed feature space \( \mathbf{X} \) for $K$ clusters, we generate $\frac{N}{K}$ samples per cluster.
For each sample \( i \in \{1, \dots, N\} \), the feature variables \( \mathbf{X}_i \in \mathbb{R}^D \) follow the same Gaussian distribution \((\mu = 1, \sigma = 2)\) across all clusters.  
We focus solely on generating the challenging case of non-separable \( \mathbf{X} \) 
since CLR already performs well in separable cases (cf.  Tab.~\ref{tab:four_cases}). We define \( K \) sets of coefficient $\mathbf{w}$, with \( \mathbf{w}_k \) serving as the ground truth for cluster \( k \). The target variable \( \mathbf{y}_i \) then is computed based on \( \mathbf{w}_k \), \( \mathbf{X}_i \), and Gaussian noise.

Following previous work on LPC~\citep{bertsimas2007classification, chembu2023generalized, zhu2012clusterwise}, we focus on cases where $K = 2,3$. SD1 consists of \( K = 2 \) clusters where the two regression lines are orthogonal. SD2 consists of \( K = 3 \) regression lines that represent positive, negative, and uncorrelated relationships, as illustrated in Fig.~\ref{fig:illustration_of_clr} (cf.  App.~C for detail setup). 

We need to avoid using a large number of clusters $K$ since for two key reasons: (1) 
GlobalOpt becomes computationally infeasible for large K values using available MIP solvers, and (2) as $K$ increases, 
multiple distinct cluster assignments may yield the same error for the defined clustering objective, leading to spurious misalignments with ground truth labels. Our experimental setup ensures each cluster reflects an identifiable and examinable linear relationship under the LPC objective. 
In App.~F.2, we further explain the rationale of this design choice and empirically verify these behaviors in the setting with $K = 5$.

\paragraph{Real-world Datasets} We selected 10 datasets from the UCI Machine Learning repository~\citep{kelly2023uci}. Each dataset was partitioned into \( K \) clusters based on the value of its categorical feature, and coefficients $\textbf{w}_k$ were fitted on samples in each cluster separately.

We ensured each dataset had at least two clusters suitable for LPC. Specifically, we identified the two clusters with the largest $l_2$ distance in their fitted linear models, defined by \( \max_{i,j} \|\mathbf{w}_i - \mathbf{w}_j\|_2 \) for \( i, j \in \{1, \dots, K\} \). App.~G further analyzes the $l_2$ distance across all possible pairs of clusters and finds that, in most datasets, only two clusters exhibit significant differences in their fitted linear models. For datasets with multiple categorical features, the feature with the largest \( \|\mathbf{w}_i - \mathbf{w}_j\|_2 \) was chosen. Datasets also required a lower cluster-wise SSE than standard regression.

Finally, the selected groups were merged to construct the final dataset, with the categorical variable used for clustering removed to ensure that the model learns solely from the feature variables. We refer the reader to App.~G for further details on each dataset and preprocessing steps.

\paragraph{Implementation} GlobalOpt follows the formulation in \eqref{eq:clr_miqp_sse}.  Greedy optimization follows the algorithm proposed by~\citet{chembu2023generalized}. Since Greedy is sensitive to initialization, each experimental trial evaluates Greedy Optimization using 20 different initializations with a maximum of 100 iterations and reports the mean performance.
We used the industry-standard solver \texttt{Gurobi} 
\citep{gurobi}, with an optimality gap fixed at 5\%. Although LPC-NS-QPBO is formulated as a QPBO problem, we observed that the MIP solver in Gurobi outperforms the PBO solver SCIP
\citep{BolusaniEtal2024ZR} (cf. App.~E for performance evaluations). We rely on Gurobi’s $M$ optimization and were unable to outperform it using customized constraints.

\subsection{Evaluation Metrics}
\paragraph{Difference From Optimal Objective}
Let $O$ represent the globally optimized LPC objective obtained from GlobalOpt. Let  $O^\prime$  denote the optimal LPC objective achieved by each method. For LPC-NS, $O^\prime$ is recalculated using \( \mathbf{w}^{\text{refit}} \) and \( \mathbf{Z} \) instead of the optimal objective given by constrained optimization using approximated $\mathbf{w}^\ast$. $O^\prime - O$ quantifies the deviation of each method from GlobalOpt. 

\paragraph{Cluster Label Difference}
Let \( \mathbf{Z}^{\text{gt}} \in \mathbb{R}^{K \times N} \) denote the binary matrix representing the ground truth cluster assignments, and let \( \mathbf{Z'} \in \mathbb{R}^{K \times N} \) be the predicted cluster assignments obtained using LPC and aligned with $\mathbf{Z'}$ using Hungarian algorithm. The discrepancy between \( \mathbf{Z}^\text{gt} \) and \( \mathbf{Z'} \) is defined as the percentage of samples with misassigned cluster labels across all \( K \) clusters:
$\text{mismatch}(\mathbf{Z}^{\text{gt}}, \mathbf{Z}^\prime) = \frac{1}{N} \sum_{k=1}^K \|\mathbf{Z}^{\text{gt}}_k - \mathbf{Z}^\prime_k\|_1$.
It represents the percentage of misassigned samples relative to the total number of samples.

\paragraph{Regression Coefficient Difference} Let \( \mathbf{w}^{\text{gt}}\in \mathbb{R}^{K \times (D + 1)} \) be the clusterwise regression coefficients based on $\mathbf{Z}^{\text{gt}}$, and let \( \mathbf{w}^\prime \in \mathbb{R}^{K \times (D + 1)} \) denote the predicted regression coefficients obtained by different methods. 
The error between \( \mathbf{w}^{\text{gt}} \) and \( \mathbf{w}^\prime \) is measured as:
$\text{difference}(\mathbf{w}^{\text{gt}}, \mathbf{w}^\prime) = \sum_{k=1}^K \|\mathbf{w}^{\text{gt}}_{k} - \mathbf{w}^\prime_{k}\|_2.$

\section{Experimental Results}

\begin{figure}
    \centering
    \includegraphics[width=\linewidth]{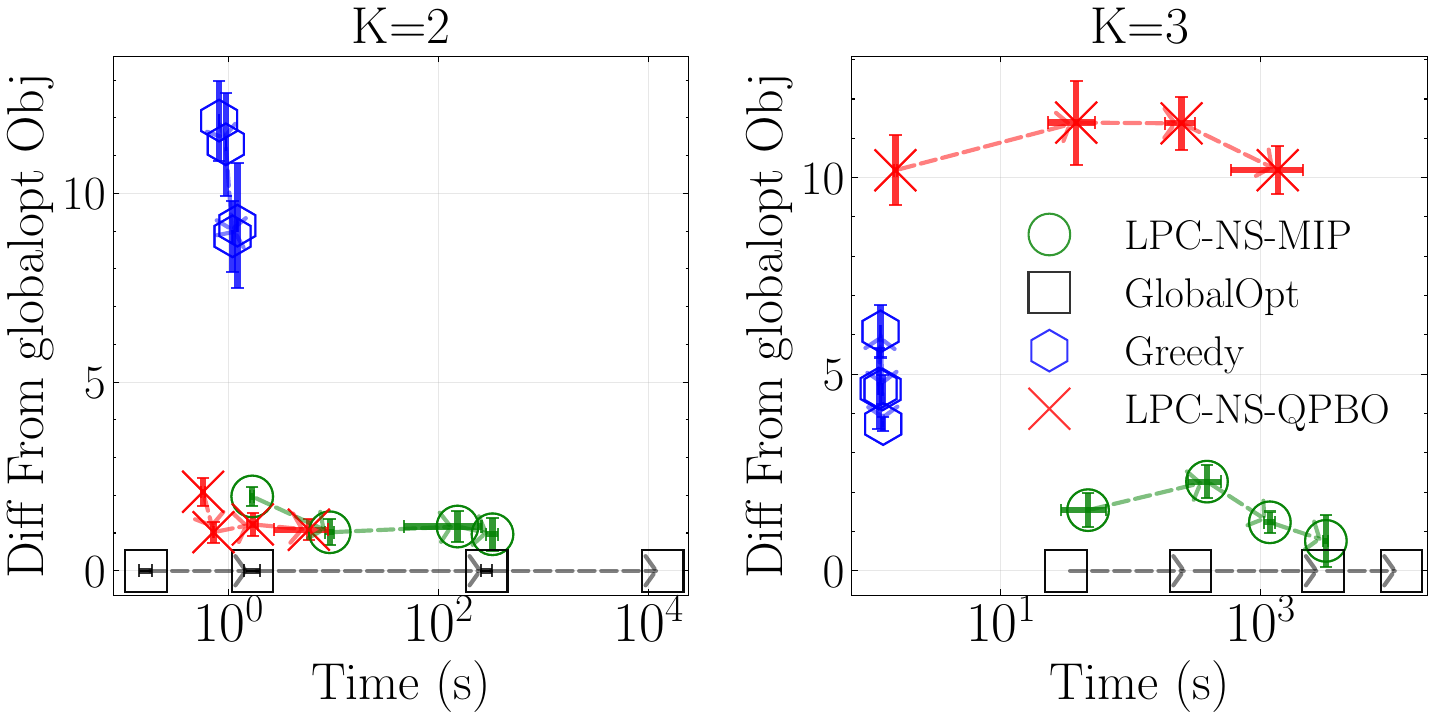}
    \caption{\textit{RQ1} Trade-off between difference from the GlobalOpt objective (y-axis) and optimization time (seconds) (x-axis) across different methods as sample sizes increase. The sample size is limited to 200 for \( K = 2 \) and 90 for \( K = 3 \), to allow GlobalOpt to achieve optimality in $\leq 2$ hours (cf. Fig.~\ref{fig:time_n}). Each data point represents the mean over trials and the solid interval indicates the 95\% confidence intervals.}
    \label{fig:RQ1_tradeoff}
\end{figure}

\paragraph{RQ1} Fig.~\ref{fig:RQ1_tradeoff} illustrates the trade-off between deviation from GlobalOpt and computational time across $K = 2$ (sample range $\{ 50,\ldots,200 \}$) and $K =  3$ (sample range $\{ 40,\ldots,60 \}$) in a high noise setting ($\sigma = 3.5$).
\emph{Sample sizes were limited for GlobalOpt to run to optimality in $\leq 2$ hours\footnote{Runtime from the start to the completion of the \texttt{Gurobi} solver with an Intel(R) Xeon(R) Platinum 8260 CPU 32 threads.}.}

\begin{figure}
    \centering
    \includegraphics[width=1\linewidth]{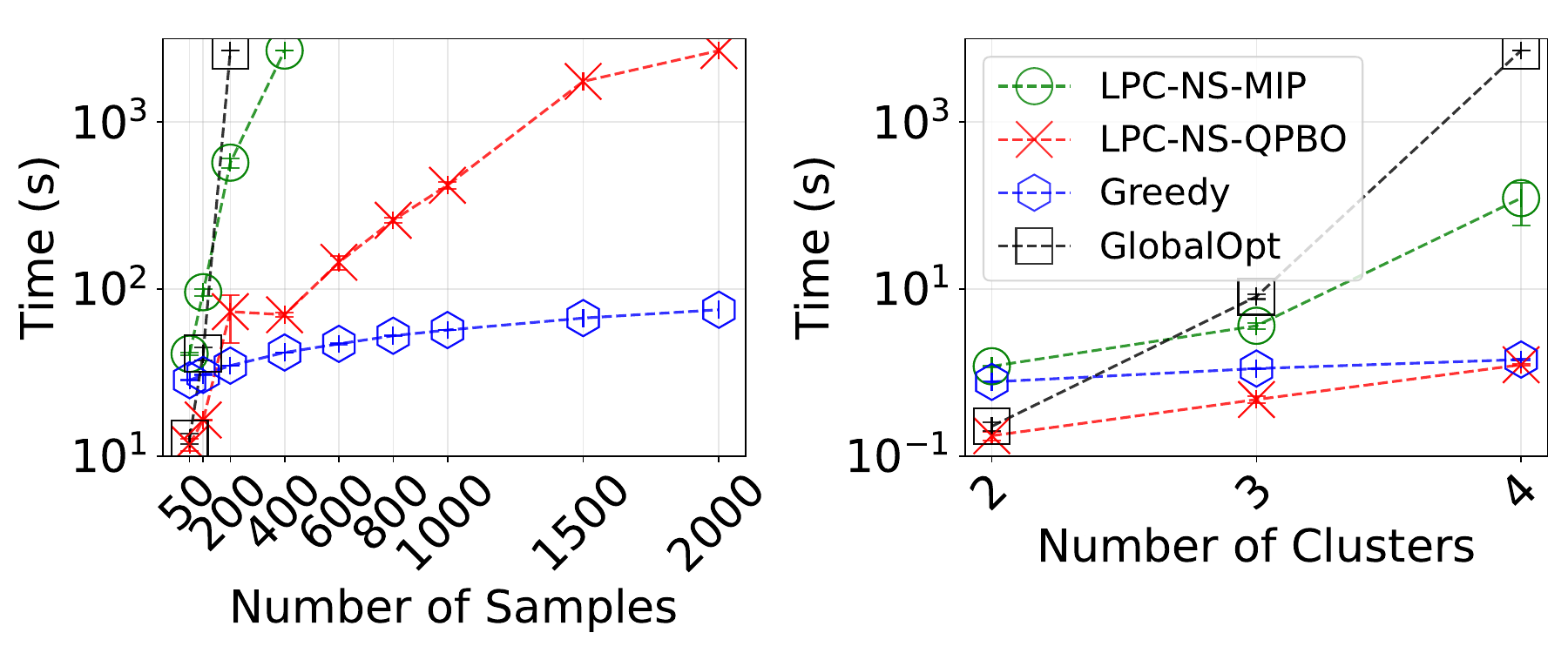}
   \caption{Runtime as the number of samples increases for \( K = 2 \) (left) and as the number of clusters increases (right). LPC-NS-QPBO exhibits better scalability with \( N = 2000 \) samples, compared to \( N = 200 \) in GlobalOpt under 2 hours.}
    \label{fig:time_n}
\end{figure}

Greedy achieves the shortest optimization time but deviates significantly from the GlobalOpt objective. In contrast, LPC-NS-QPBO maintains a comparable runtime to Greedy for smaller sample sizes while achieving a substantially lower deviation, indicating a near-optimal solution for all samples in \( K = 2 \) but exhibiting higher error for \( K = 3 \). (Fig.~\ref{fig:time_n}-left) further shows that LPC-NS-QPBO optimizes 2000 samples in \( K = 2 \) within the same time limit, whereas GlobalOpt reaches its limit at 200 samples. Scalability declined significantly for all methods in \( K > 2 \) (Fig.~\ref{fig:time_n}-right), as samples exhibited less distinct relationships across clusters. Although LPC-NS-MIP requires a longer optimization time, it consistently achieves a more optimized solution than Greedy in both cases while remaining significantly more efficient than GlobalOpt. GlobalOpt attains the optimal objective value but at a higher computational cost, with the efficiency gap widening as the number of samples increases.

\begin{figure}[!h]
    \centering
    \includegraphics[width=\linewidth]{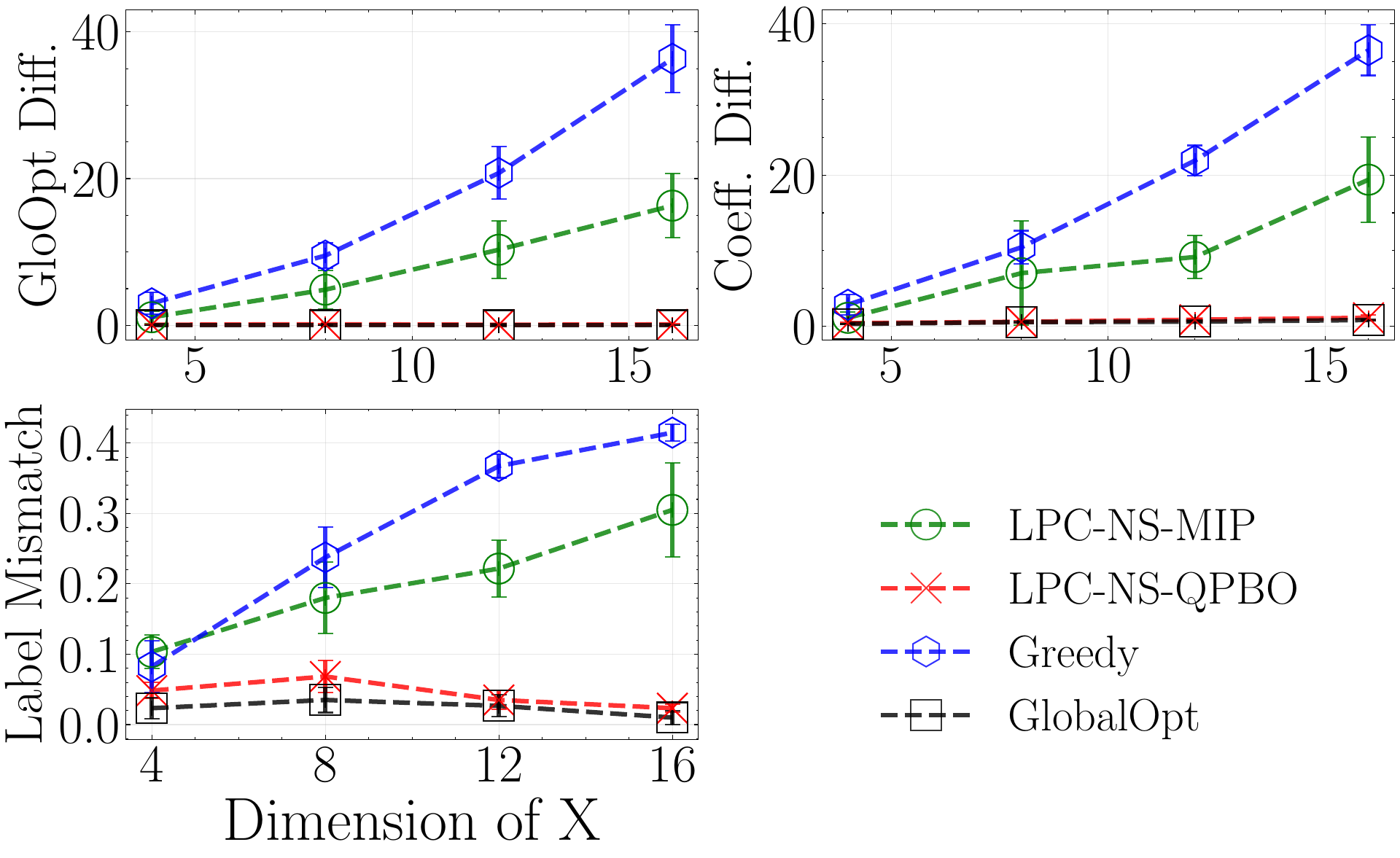   } 
\caption{\textit{RQ2 Feature Variables}  The performance difference across methods as the number of feature variables increases.}
    \label{fig:D}
\end{figure}

\begin{figure}
    \centering
    \includegraphics[width=0.9\linewidth]{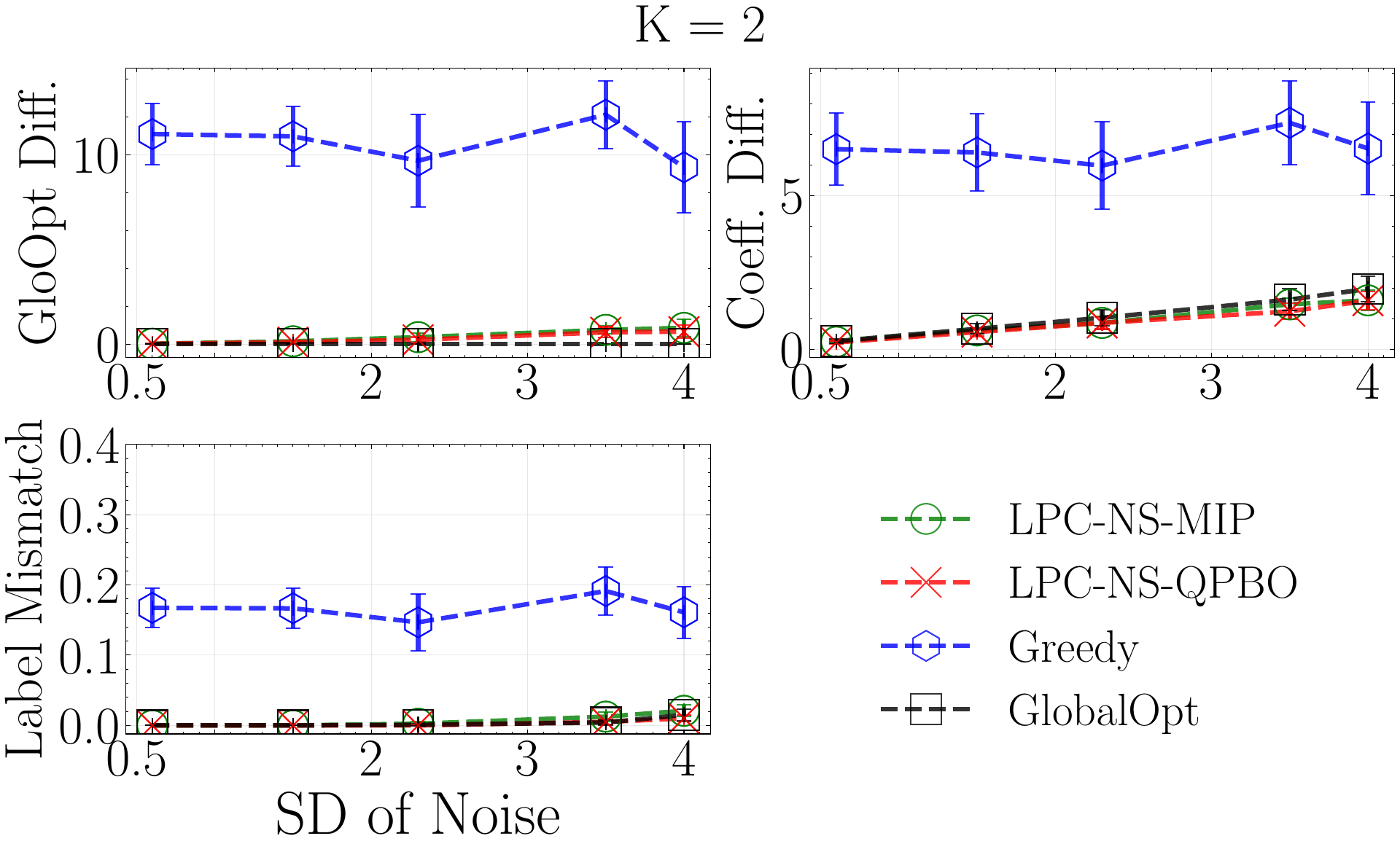}
    \includegraphics[width=0.9\linewidth]{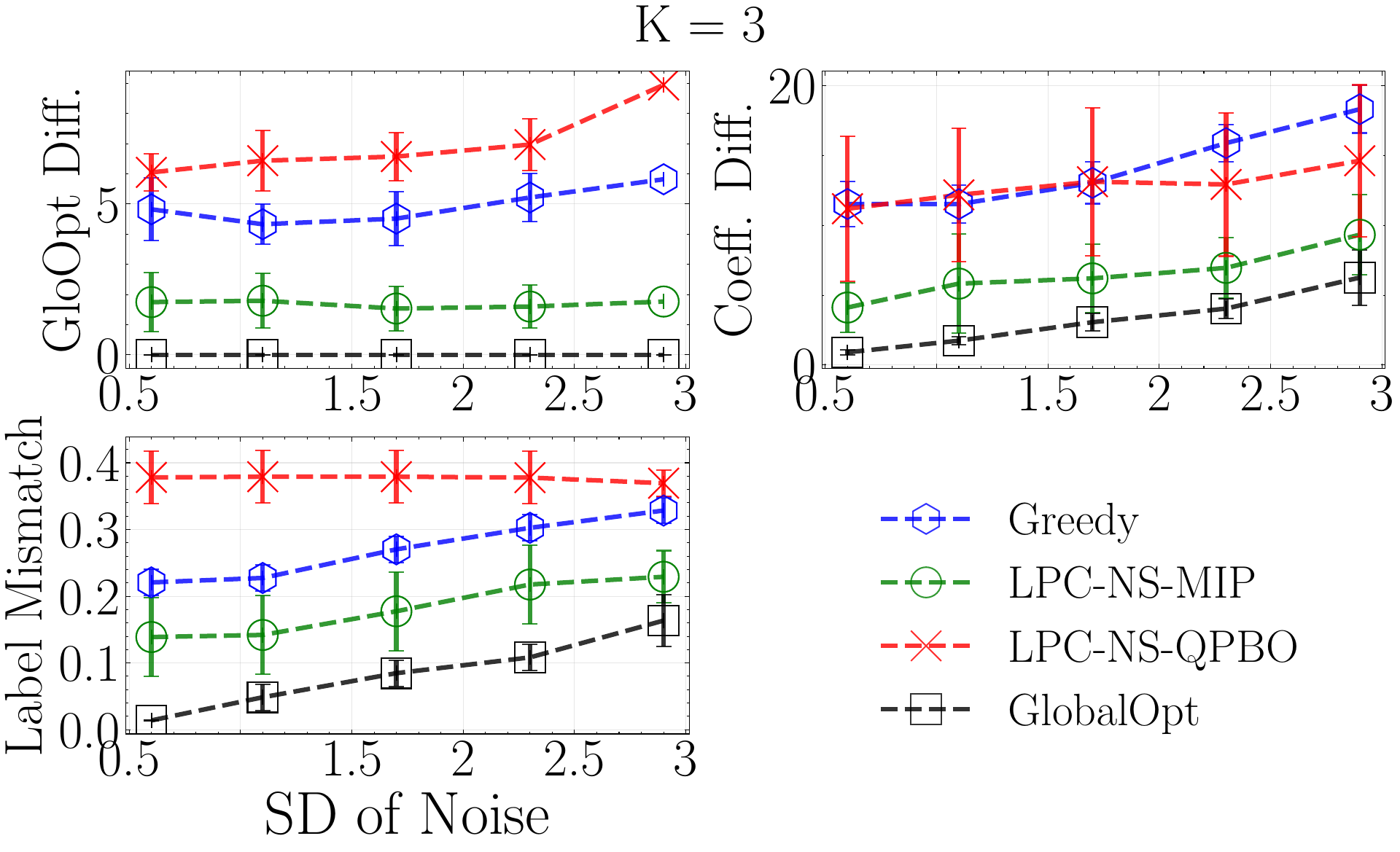}
\caption{\textit{RQ2 Noise} The performance difference across different methods in an increasing level of Gaussian noise.}
    \label{fig:noise}
\end{figure}

\begin{figure}
    \centering
    \includegraphics[width=0.85\linewidth]{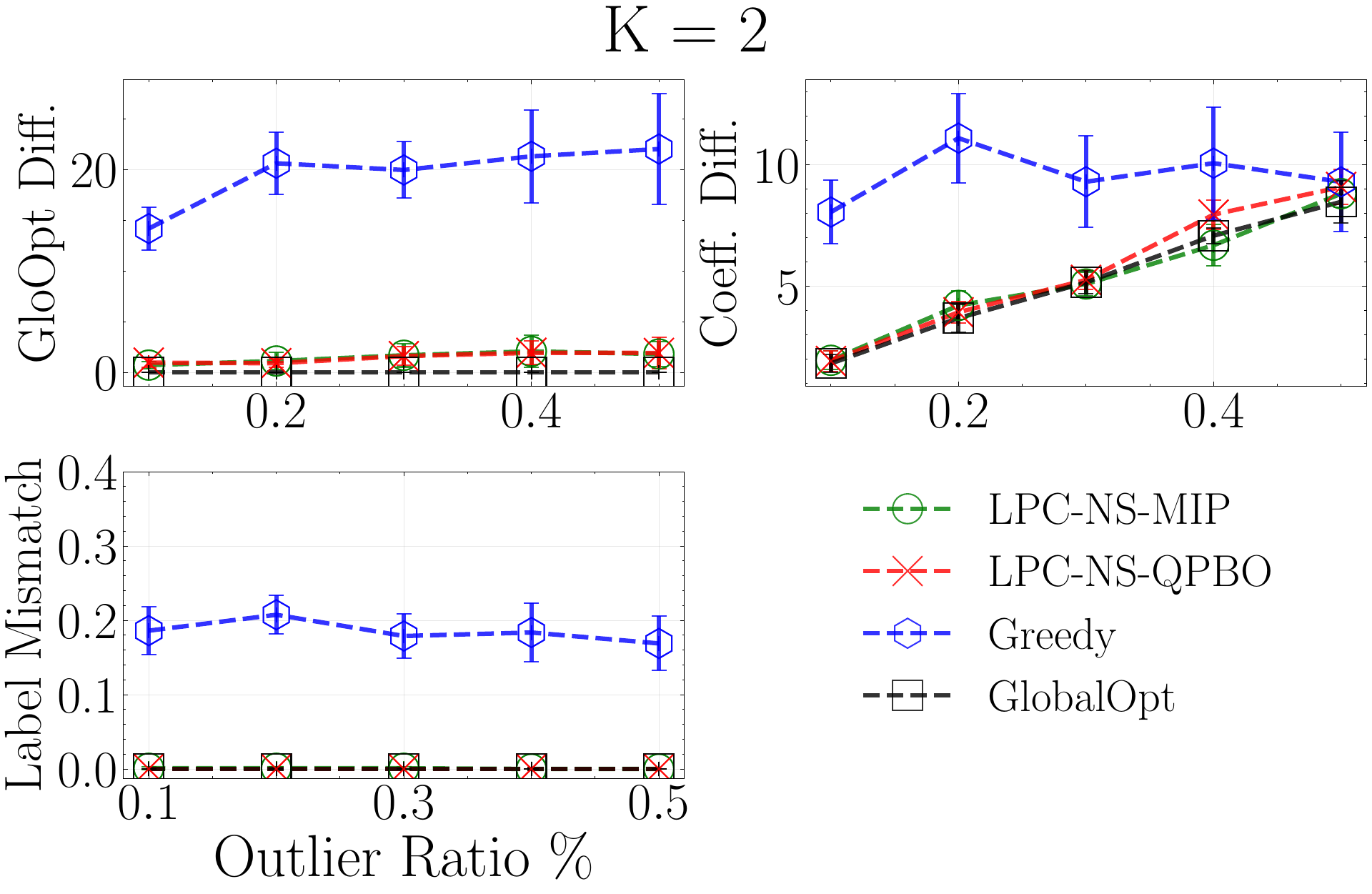}
\caption{\textit{RQ2 Outliers}  The performance of different methods as the proportion of outliers in target variables increases.}
    \label{fig:Outlier}
\end{figure}

\paragraph{RQ2} We evaluate deviations from the GlobalOpt objective and the performance of recovering \( \mathbf{w}^{\text{gt}} \) and \( \mathbf{Z}^{\text{gt}} \) across different methods under varying conditions. We analyze the impact of varying the level of Gaussian noise in the target variable (Fig.~\ref{fig:noise}), feature dimensionality (Fig.~\ref{fig:D}), and the proportion of outliers (Fig.~\ref{fig:Outlier})
in the target variable. 
LPC-NS-MIP consistently outperforms Greedy for both \( K = 2 \) and \( K = 3 \) (cf. Appendix F), achieving a significantly smaller gap from the globally optimal objective. LPC-NS-QPBO also outperforms Greedy for \( K = 2 \) but performs suboptimally for \( K = 3 \). Additionally, both proposed methods achieve higher accuracy in recovering ground truth $\mathbf{Z}$ and $\mathbf{w}$ for \( K = 2 \), while LPC-NS-MIP maintains its superiority in \( K = 3 \).

The performance of Greedy deteriorates in high-dimensional settings (Fig.~\ref{fig:D}), whereas LPC-NS-QPBO remains robust, maintaining near-optimal objective values. In addition, LPC-NS-MIP continues to outperform Greedy as the proportion of outliers increases.

\begin{table*}
    \centering
    \small
    \begin{tabular}{c|rrr|c} 
        \hline
        \multicolumn{1}{c|}{\textbf{Dataset}} & \multicolumn{3}{c|}{\textbf{SSE}} & \textbf{Separability} \\
        Dataset (UCI Repository ID) & LPC-NS-QPBO & LPC-NS-MIP & Greedy & $  \mathcal{E}_{\text{sep}} $ \\
        \hline
        Stock Portfolio Performance (390) & 16.1798 & \textbf{12.4745} & $17.1278 \pm 1.6503$ & 0.0000 \\
        
        Servo (87) & 12.4865 & \textbf{2.5305} 
        & $4.0431 \pm 1.1435$ & 0.0543 \\
        
        Solar Flare (89) & \textbf{3.8289} & 11.0511 & $8.3014 \pm 1.042$ & 0.5885\\ 
        
        Productivity Prediction (597) & 41.4977 & \textbf{38.2234} & $43.4662 \pm 2.7841$ & 0.5914\\
        
          Heart Failure Prediction (519) & \( 4.733 \times 10^{-30} \) & \textbf{\( \mathbf{6.920 \times 10^{-32} } \)} & $25.0644 \pm 8.6361 $ & 0.7685\\

       Liver Disorders (60) & 108.3983 & \textbf{94.9041} & $112.0968 \pm 5.4914$ & 0.9784\\

       Student Performance (320) &  \textbf{4.8812} &5.6557& \( 6.8896 \pm 0.4574\) & 2.0107\\
               
        Parkinsons Telemonitoring (189) & \textbf{135.3372} &	146.9704 &	$ 174.6422 \pm 12.7622$ & 3.9580 \\	

         Facebook Metrics (368) & \( 2.882 \times 10^{-20} \) & \textbf{\( \mathbf{2.0464 \times 10^{-20}} \)} & \( 3.1 \times 10^{-5} \pm 1.2 \times 10^{-5}  \) & 5.6163\\
         
        Infrared Thermography (925) & \( 1.080 \times 10^{-13} \) & \( \mathbf{6.904 \times 10^{-15}} \) & $0.1029 \pm 0.038$ & 7.9983\\
        \hline
    \end{tabular}
        \caption{\textit{RQ3} Performance across different methods in 10 real datasets. GlobalOpt is excluded due to infeasible runtime on this scale. Mean and 95\% confidence interval in Greedy (20 initializations) is reported. }
    \label{real_data}
\end{table*}

\paragraph{RQ3} To assess performance in real-world scenarios, Tab.~\ref{real_data} and App.~G.3 report the performance of different methods\footnote{Due to scalability constraints, GlobalOpt was not evaluated as it is infeasible for most datasets within a reasonable runtime.} across 10 real-world datasets, along with the separability measure \( \mathcal{E}_{\text{sep}} \). LPC-NS-MIP and LPC-NS-QPBO outperform Greedy. In some separable datasets, LPC-NS surpasses Greedy in cases where relationships are easier to identify (low SSE), where LPC-NS is more robust to noise, or when clusters are highly imbalanced -- scenarios in which Greedy fails to recover correct partitions. 

We also observed that the performance of LPC-NS-MIP and LPC-NS-QBPO is highly dependent on the dataset. If scalability is a concern, LPC-NS-QBPO is the preferred choice, as it significantly improves scalability with comparable performance in the majority of cases.

\section{Conclusion}  
With the aim of developing scalable and near-optimal LPC optimization methods, we leverage properties of non-separable feature spaces to derive novel MIP and QPBO reductions from an existing globally optimal MIP formulation. Specifically, LPC-NS-MIP offers a more compact and efficient formulation than GlobalOpt while maintaining near-optimal performance in non-separable settings. LPC-NS-QPBO further enhances scalability via a QPBO reduction. Extensive experiments show that LPC-NS-MIP consistently outperforms Greedy in optimizing the objective and recovering hidden cluster assignments and regression coefficients across varying noise, dimensions, and outliers, while LPC-NS-QPBO remains highly competitive for \(K = 2\).

\bibliography{reference}

\newpage
\onecolumn
\appendix
\section{Error analysis between optimal regression coefficients $\mathbf{w}$ and the $\mathbf{w}^\ast$ approximation} \label{approximation_error}
 
Given $k$ partition of samples in target variable $\mathbf{y} = \begin{bmatrix}\mathbf{y}_k \\ \dots \\\mathbf{y}_K \end{bmatrix}$ and feature variables $\mathbf{X} = \begin{bmatrix}\mathbf{X}_1 \\  \dots \\\mathbf{X}_K \end{bmatrix}$. The objective of each cluster $k \in \{1,..,K\}$ is given as
 \[ 
O_k = \| \mathbf{y}_k - \mathbf{X}_k \mathbf{w}_k \|_2^2 + \lambda\|\mathbf{w}_k\|_2^2
 \]
 where 
 \[
\mathbf{w}_k = (\mathbf{X}^\top\mathbf{Z}_k\mathbf{X} + \lambda  \mathbf{I})^{-1} \mathbf{X}_k^\top \mathbf{y}_k
\]
We approximated $\mathbf{w}_k$ as $\mathbf{w}_k^\ast = (\alpha_k\mathbf{X}^\top\mathbf{X} + \lambda  \mathbf{I})^{-1}\mathbf{X}_k\mathbf{y}_k$. The approximated loss function will be
 \[ 
O_k^\ast = \| \mathbf{y}_k - \mathbf{X}_k \mathbf{w}^\ast_k \|_2^2 + \lambda\|\mathbf{w}^\ast_k\|_2^2
 \]
where 
\[
\mathbf{w}^\ast_k = \left(  \alpha_k\mathbf{X}^\top\mathbf{X} + \lambda  \mathbf{I} \right)^{-1} \mathbf{X}_k^\top \mathbf{y}_k
\]
 
For the rest of the deviation, we assume a balanced cluster size ($N/K$), which suggests $\alpha = \frac{1}{K}$. For the imbalanced cluster size setting, please refer to App.~D for an empirical evaluation of the performance of the proposed methods using iteratively tuned $\alpha$. Under this assumption, the sum of the differences between all representations is the approximation error introduced by our proposed method in comparison to the original globally optimal objective. 

In specific,
 \begin{equation}\label{difference_in_objecitve}
      \sum_{k=1}^K \left( O_k^\ast - O_k \right)
 \end{equation}
To simplify  \eqref{difference_in_objecitve}, we first simplified $O_k$ and $O_k^\ast$ from their norm forms.
 \begin{align}
O_k &= \| \mathbf{y}_k - \mathbf{X}_k (\mathbf{X}^\top\mathbf{Z}_k\mathbf{X} + \lambda  \mathbf{I})^{-1} \mathbf{X}_k^\top \mathbf{y}_k \|_2^2 + \lambda\|(\mathbf{X}^\top\mathbf{Z}_k\mathbf{X} + \lambda  \mathbf{I})^{-1} \mathbf{X}_k^\top \mathbf{y}_k \|_2^2 \\
&=\|\mathbf{R_1} \mathbf{y}_k \|_2^2 + \lambda\|(\mathbf{X}^\top\mathbf{Z}_k\mathbf{X} + \lambda  \mathbf{I})^{-1} \mathbf{X}_k\mathbf{y}_k \|_2^2
 \end{align}
where $\mathbf{R_1} = I - \mathbf{X}_k (\mathbf{X}^\top\mathbf{Z}_k\mathbf{X} + \lambda  \mathbf{I})^{-1} \mathbf{X}_k^\top  $. $\mathbf{R_1}$. which are idempotent. Thus,  $\mathbf{R_1}^2 = \mathbf{R_1}$
  \begin{align}
O_k &= \mathbf{y}_k^\top \mathbf{R_1} \mathbf{y}_k + \lambda((\mathbf{X}^\top\mathbf{Z}_k\mathbf{X} + \lambda  \mathbf{I})^{-1} \mathbf{X}_k^\top \mathbf{y}_k)^\top((\mathbf{X}^\top\mathbf{Z}_k\mathbf{X} + \lambda  \mathbf{I})^{-1} \mathbf{X}_k^\top \mathbf{y}_k) \\
&= \mathbf{y}_k^\top \mathbf{R_1} \mathbf{y}_k + \lambda \mathbf{y}_k^\top \mathbf{X}_k(\mathbf{X}^\top\mathbf{Z}_k\mathbf{X} + \lambda  \mathbf{I})^{-1\top}(\mathbf{X}^\top\mathbf{Z}_k\mathbf{X} + \lambda  \mathbf{I})^{-1} \mathbf{X}_k^\top \mathbf{y}_k 
 \end{align} 
 Since $(\mathbf{X}^\top\mathbf{Z}_k\mathbf{X} + \lambda  \mathbf{I})^{-1}$ is symmetric, 
\begin{align}
O_k &= \mathbf{y}_k^\top \mathbf{R_1} \mathbf{y}_k + \lambda \mathbf{y}_k^\top \mathbf{X}_k (\mathbf{X}^\top\mathbf{Z}_k\mathbf{X} + \lambda  \mathbf{I})^{-2} \mathbf{X}_k^\top \mathbf{y}_k  \label{Ok}
 \end{align} 
Similarly,
 \begin{align}
O_k^\ast &= \| \mathbf{y}_k - \mathbf{X}_k (\alpha_k\mathbf{X}^\top\mathbf{X} + \lambda  \mathbf{I})^{-1} \mathbf{X}_k^\top \mathbf{y}_k \|_2^2 + \lambda\|(\alpha_k\mathbf{X}^\top\mathbf{X} + \lambda  \mathbf{I})^{-1} \mathbf{X}_k^\top \mathbf{y}_k \|_2^2 \\
&=  \mathbf{y}_k^\top \mathbf{R_1}^\ast \mathbf{y}_k  + \lambda \mathbf{y}_k^\top \mathbf{X}_k (\alpha_k\mathbf{X}^\top\mathbf{X}+ \lambda  \mathbf{I})^{-2} \mathbf{X}_k^\top \mathbf{y}_k \label{Okast}
 \end{align} 
where $\mathbf{R_1}^\ast = I - \mathbf{X}_k (\alpha_k\mathbf{X}^\top\mathbf{X} + \lambda  \mathbf{I})^{-1} \mathbf{X}_k^\top$.

Combining \eqref{Ok} and \eqref{Okast}, the difference between $O^\ast$ and $O_k^\ast$ will be
\begin{align}
O_k^\ast - O_k &= \mathbf{y}_k^\top \mathbf{R_1}^\ast \mathbf{y}_k  + \lambda \mathbf{y}_k^\top \mathbf{X}_k (\alpha_k\mathbf{X}^\top\mathbf{X}+ \lambda  \mathbf{I})^{-2} \mathbf{X}_k^\top \mathbf{y}_k 
 - \mathbf{y}_k^\top \mathbf{R_1} \mathbf{y}_k - \lambda \mathbf{y}_k^\top \mathbf{X}_k \nonumber \\
& \quad \quad (\mathbf{X}^\top\mathbf{Z}_k\mathbf{X} + \lambda  \mathbf{I})^{-2} \mathbf{X}_k^\top \mathbf{y}_k \\
&= \mathbf{y}_k^\top (\mathbf{R_1}^\ast - \mathbf{R_1}) \mathbf{y}_k  + \lambda \mathbf{y}_k^\top \mathbf{X}_k((\alpha_k\mathbf{X}^\top\mathbf{X} + \lambda  \mathbf{I})^{-2} - (\mathbf{X}^\top\mathbf{Z}_k\mathbf{X} + \lambda  \mathbf{I})^{-2}) \mathbf{X}_k^\top \mathbf{y}_k \label{difference_OK_OKAST}
 \end{align} 

For a scalar value \( j \), the L2 norm is given by \( \|j\|_2 = |j| \geq j \). Consequently, the upper bound of the approximation error, \( O_k^\ast - O_k \), is also bounded above by \( |O_k^\ast - O_k| = |O_k - O_k^\ast| = \|O_k^\ast - O_k^\ast\|_2 \). Using  \eqref{difference_OK_OKAST}, the following inequality holds:

\begin{equation} 
    O_k^\ast - O_k \leq \|\mathbf{y}_k^\top (\mathbf{R_1} - \mathbf{R_1}^\ast) \mathbf{y}_k\|_2 
    + \| \lambda \mathbf{y}_k^\top \mathbf{X}_k 
    \left( (\mathbf{X}_k^\top \mathbf{X}_k + \lambda \mathbf{I})^{-2} 
    - (\alpha_k\mathbf{X}^\top\mathbf{X} + \lambda \mathbf{I})^{-2} \right) 
    \mathbf{X}_k^\top \mathbf{y}_k\|_2. \label{upper_bound}
\end{equation}

In  \eqref{upper_bound}, the first term $ \|\mathbf{y}_k^\top (\mathbf{R_1} - \mathbf{R_1}^\ast) \mathbf{y}_k \|_2$:
 \begin{align}
\|\mathbf{y}_k^\top (\mathbf{R_1} - \mathbf{R_1}^\ast) \mathbf{y}_k\|_2
 &= \|\mathbf{y}_k^\top (\mathbf{X}_k (\alpha_k\mathbf{X}^\top\mathbf{X} + \lambda  \mathbf{I})^{-1} \mathbf{X}_k^\top - \mathbf{X}_k (\mathbf{X}^\top\mathbf{Z}_k\mathbf{X} + \lambda  \mathbf{I})^{-1} \mathbf{X}_k^\top) \mathbf{y}_k\|_2\\
 &=  \|\mathbf{y}_k^\top \mathbf{X}_k ((\alpha_k\mathbf{X}^\top\mathbf{X} + \lambda  \mathbf{I})^{-1} - (\mathbf{X}^\top\mathbf{Z}_k\mathbf{X} + \lambda  \mathbf{I})^{-1} ) \mathbf{X}_k^\top \mathbf{y}_k\|_2\\
  &=  \|\mathbf{y}_k^\top \mathbf{X}_k (\alpha_k\mathbf{X}^\top\mathbf{X} + \lambda  \mathbf{I})^{-1} (\mathbf{X}^\top\mathbf{Z}_k\mathbf{X} - \alpha_k\mathbf{X}^\top\mathbf{X})(\mathbf{X}^\top\mathbf{Z}_k\mathbf{X} + \lambda  \mathbf{I})^{-1} \mathbf{X}_k^\top \mathbf{y}_k\|_2 
   \end{align}
Based on the submultiplicative property of a matrix,
    \begin{align}
  \|\mathbf{y}_k^\top (\mathbf{R_1} - \mathbf{R_1}^\ast) \mathbf{y}_k\|_2 &\leq \|\mathbf{y}_k^\top \mathbf{X}_k\|_2 \| (\alpha_k\mathbf{X}^\top\mathbf{X} + \lambda  \mathbf{I})^{-1}\|_2 \| (\mathbf{X}^\top\mathbf{Z}_k\mathbf{X} - \alpha_k\mathbf{X}^\top\mathbf{X})\|_2 \nonumber \\ &
\quad \quad \|(\mathbf{X}^\top\mathbf{Z}_k\mathbf{X} + \lambda  \mathbf{I})^{-1} \|_2 \|\mathbf{X}_k\mathbf{y}_k\|_2 
    \end{align}
Since\ $\|\mathbf{A}\mathbf{v}\|_2 \leq \|\mathbf{A}\|_2\|\mathbf{v}\|_2$ for a matrix $\mathbf{A}$ and a vector $\mathbf{v}$,
    \begin{align}
\|\mathbf{y}_k^\top (\mathbf{R_1} - \mathbf{R_1}^\ast) \mathbf{y}_k\|_2 &\leq \|\mathbf{y}_k\|_2 \| \mathbf{X}_k\|_2 \| (\alpha_k\mathbf{X}^\top\mathbf{X} + \lambda  \mathbf{I})^{-1}\|_2 (\mathbf{X}^\top\mathbf{Z}_k\mathbf{X} - \alpha_k\mathbf{X}^\top\mathbf{X})\|_2\|  \nonumber \\
& \quad \quad \| (\mathbf{X}^\top\mathbf{Z}_k\mathbf{X} + \lambda  \mathbf{I})^{-1} \|_2 \|\mathbf{X}_k\|_2 \| \mathbf{y}_k\|_2 \\
&\leq \|\mathbf{y}_k\|^2_2 \| \mathbf{X}_k\|^2_2 \| (\alpha_k\mathbf{X}^\top\mathbf{X} + \lambda  \mathbf{I})^{-1}\|_2  \nonumber \\
&\quad \quad\| (\mathbf{X}^\top\mathbf{Z}_k\mathbf{X} - \alpha_k\mathbf{X}^\top\mathbf{X})\|_2\|(\mathbf{X}^\top\mathbf{Z}_k\mathbf{X} + \lambda  \mathbf{I})^{-1} \|_2 
 \end{align}
The regularization parameter \( \lambda \) is a positive scalar. Since \( \alpha_k\mathbf{X}^\top\mathbf{X} \), the uncentered covariance matrix, is always positive semidefinite, the minimum singular value of \( (\alpha_k\mathbf{X}^\top\mathbf{X} + \lambda \mathbf{I}) \), denoted as \( \sigma_{\min}(\alpha_k\mathbf{X}^\top\mathbf{X} + \lambda \mathbf{I}) \), satisfies:
\[
\sigma_{\min}(\alpha_k\mathbf{X}^\top\mathbf{X} + \lambda \mathbf{I}) \geq \lambda.
\]
Consequently, the maximum singular value of \( (\alpha_k\mathbf{X}^\top\mathbf{X} + \lambda \mathbf{I})^{-1} \), denoted as \( \sigma_{\max}((\alpha_k\mathbf{X}^\top\mathbf{X} + \lambda \mathbf{I})^{-1}) \), is upper-bounded by:
\[
\sigma_{\max}((\alpha_k\mathbf{X}^\top\mathbf{X} + \lambda \mathbf{I})^{-1}) \leq \frac{1}{\lambda}.
\]
  \begin{align}
     \|\mathbf{y}_k^\top (\mathbf{R_1} - \mathbf{R_1}^\ast) \mathbf{y}_k\|_2 
                    &\leq \frac{\|\mathbf{y}_k\|^2_2 \| \mathbf{X}_k\|^2_2 \| (\mathbf{X}^\top\mathbf{Z}_k\mathbf{X} - \alpha_k\mathbf{X}^\top\mathbf{X})\|_2}{\lambda^2}  
 \end{align}

 In  \eqref{upper_bound}, the second term 
 \begin{align}
& \|\lambda \mathbf{y}_k^\top \mathbf{X}_k((\mathbf{X}^\top\mathbf{Z}_k\mathbf{X} + \lambda  \mathbf{I})^{-2} - (\alpha_k\mathbf{X}^\top\mathbf{X} + \lambda  \mathbf{I})^{-2}) \mathbf{X}_k^\top \mathbf{y}_k\|_2 \nonumber\\
&\leq \lambda \|\mathbf{y}_k\|^2_2 \|  \mathbf{X}_k\|^2_2 \| ((\mathbf{X}^\top\mathbf{Z}_k\mathbf{X} + \lambda  \mathbf{I})^{-2} - (\alpha_k\mathbf{X}^\top\mathbf{X} + \lambda  \mathbf{I})^{-2})\|_2
\end{align}
Since 
\begin{equation}  
 (\mathbf{X}_k^\top \mathbf{X}_k + \lambda  \mathbf{I})^{-2} - (\alpha_k\mathbf{X}^\top\mathbf{X} + \lambda  \mathbf{I})^{-2} = (\mathbf{X}_k^\top \mathbf{X}_k + \lambda  \mathbf{I})^{-1} \Delta (\alpha_k\mathbf{X}^\top\mathbf{X} + \lambda  \mathbf{I})^{-1} (\mathbf{X}_k^\top \mathbf{X}_k + \alpha_k\mathbf{X}^\top\mathbf{X} + 2\lambda  \mathbf{I})^{-1}
\end{equation}
where \(\Delta = (\alpha_k\mathbf{X}^\top\mathbf{X} + \lambda  \mathbf{I}) - (\mathbf{X}_k^\top \mathbf{X}_k + \lambda  \mathbf{I}) =  \alpha_k\mathbf{X}^\top\mathbf{X}- \mathbf{X}_k^\top \mathbf{X}_k\)
\begin{align}
& \|\lambda \mathbf{y}_k^\top \mathbf{X}_k((\mathbf{X}^\top\mathbf{Z}_k\mathbf{X} + \lambda  \mathbf{I})^{-2} - (\alpha_k\mathbf{X}^\top\mathbf{X} + \lambda  \mathbf{I})^{-2}) \mathbf{X}_k^\top \mathbf{y}_k\|_2 \nonumber\\
    &\leq \lambda \| \mathbf{y}_k \|_2^2 \| \mathbf{X}_k \|_2^2 \|(\mathbf{X}_k^\top \mathbf{X}_k + \lambda  \mathbf{I})^{-1} \|_2\|\Delta\|_2\| (\alpha_k\mathbf{X}^\top\mathbf{X}+ \lambda  \mathbf{I})^{-1}\|_2\| (\mathbf{X}_k^\top \mathbf{X}_k + \alpha_k\mathbf{X}^\top\mathbf{X}+ 2\lambda  \mathbf{I})^{-1}\|_2 \\
    &\leq \lambda \| \mathbf{y}_k \|_2^2 \| \mathbf{X}_k \|_2^2 \frac{1}{\lambda} \|\Delta\|_2 \frac{1}{\lambda} \frac{1}{2\lambda} \\
    &\leq \frac{\| \mathbf{y}_k \|_2^2 \| \mathbf{X}_k \|_2^2 \|\alpha_k\mathbf{X}^\top\mathbf{X} - \mathbf{X}_k^\top \mathbf{X}_k\|_2}{2\lambda^2} 
\end{align}

Therefore,  \eqref{upper_bound} is equal to

   \begin{align}
O_k^\ast - O_k &\leq \frac{\|\mathbf{y}_k\|^2_2 \| \mathbf{X}_k\|^2_2 \| (\mathbf{X}^\top\mathbf{Z}_k\mathbf{X} - \alpha_k\mathbf{X}^\top\mathbf{X})\|_2}{\lambda^2}  +  \frac{\| \mathbf{y}_k \|_2^2 \| \mathbf{X}_k \|_2^2 \|\alpha_k\mathbf{X}^\top\mathbf{X} - \mathbf{X}_k^\top \mathbf{X}_k\|_2}{2\lambda^2}  \\
&\leq \frac{2\|\mathbf{y}_k\|^2_2 \| \mathbf{X}_k\|^2_2 \|\alpha_k\mathbf{X}^\top\mathbf{X} - \mathbf{X}^\top\mathbf{Z}_k\mathbf{X}\|_2 + \| \mathbf{y}_k \|_2^2 \| \mathbf{X}_k \|_2^2 \| \alpha_k\mathbf{X}^\top\mathbf{X} - \mathbf{X}_k^\top \mathbf{X}_k \|_2}{2\lambda^2} \\
&\leq \frac{3\|\mathbf{y}_k\|^2_2 \| \mathbf{X}_k\|^2_2 \|\alpha_k\mathbf{X}^\top\mathbf{X} - \mathbf{X}^\top\mathbf{Z}_k\mathbf{X}\|_2 }{2\lambda^2} \\
\end{align}
The sum of the difference between $O_k^\ast - O_k$ across all clusters is upper bounded by
\begin{equation} \label{eq:final_bounded}
   \sum_{k=1}^{K} \left( O_k^\ast - O_k \right) \leq \sum_{k=1}^{K} \left( \frac{3\|\mathbf{y}_k\|^2_2 \| \mathbf{X}_k\|^2_2 \|\alpha_k\mathbf{X}^\top\mathbf{X} - \mathbf{X}^\top\mathbf{Z}_k\mathbf{X}\|_2 }{2\lambda^2} \right)
\end{equation}

It is important to note that the derivation of the error bound assumes an idealized setting in which the cluster assignments coincide with the ground truth. However, the final objective in \eqref{eq:lpc_objective_z} is inherently robust to suboptimal assignments: poor clusterings necessarily induce poor cluster-wise regression coefficients, which in turn yield large within-cluster SSE. Since the LPC objective minimizes the total SSE across all clusters, such suboptimal assignments are naturally penalized and thus excluded by the optimization process.

\newpage
\section{Simplifying the LPC-NS-MIP Objective Function in \eqref{eq:lpc_objective_z}} \label{appendix:simplifying_qpbo}
This appendix provides the detailed mathematical derivation for simplifying the objective function of LPC-NS-MIP in \eqref{eq:lpc_objective_z}. The objective is given as:

\begin{equation}\label{eq:lpc_objective_z_apendix}
    \min_{\mathbf{Z}} \sum_{k=1}^K \left( \| \mathbf{Z}_k \mathbf{y} - \mathbf{Z}_k \mathbf{X} \mathbf{w}^\ast_k \|_2^2 + \lambda \| \mathbf{w}^\ast_k \|_2^2 \right)
\end{equation}
Expanding \eqref{eq:lpc_objective_z_apendix},

\begin{align}\label{eq:lpc_objective_z_expand}
    \min_{\mathbf{Z}} \sum_{k=1}^K \left[ \mathbf{y}^\top \mathbf{Z}^{k\top} \mathbf{Z}_k\mathbf{y}- 2 \mathbf{y}^\top \mathbf{Z}^{k\top} \mathbf{Z}_k \mathbf{X} \mathbf{w}_k^\ast + {\mathbf{w}_k^\ast}^\top (\mathbf{X} )^\top \mathbf{Z}^{k\top} \mathbf{Z}_k \mathbf{X}  \mathbf{w}_k^\ast + \lambda {\mathbf{w}_k^\ast}^\top \mathbf{w}_k^\ast \right].
\end{align}

From each term in \eqref{eq:lpc_objective_z_expand},

\paragraph{Term 1: \( \mathbf{y}^\top \mathbf{Z}^{k\top} \mathbf{Z}_k\mathbf{y}= \mathbf{y}^\top \mathbf{Z}_k\mathbf{y}\)} since $\mathbf{Z}_k $ is a binary diagonal matrix.

\paragraph{Term 2: \(-2 \mathbf{y}^\top \mathbf{Z}_k \mathbf{X} \mathbf{w}_k^\ast \)}

Substituting \( \mathbf{w}_k^\ast = A^\ast (\mathbf{X})^\top \mathbf{Z}_k \mathbf{y}\):
\begin{align}
    -2\mathbf{y}^\top \mathbf{Z}_k \mathbf{X} \mathbf{w}_k^\ast &=  -2\mathbf{y}^\top \mathbf{Z}_k \mathbf{X} A^\ast \mathbf{X}^{\top} \mathbf{Z}_k\mathbf{y}
\end{align}
\paragraph{Term 3: 
\( {\mathbf{w}_k^\ast}^\top \mathbf{X}^{\top}  \mathbf{Z}_k \mathbf{X} \mathbf{w}_k^\ast \)}
\begin{align}
    {\mathbf{w}_k^\ast}^\top \mathbf{X}^{\top}  \mathbf{Z}_k \mathbf{X} \mathbf{w}_k^\ast &= \left( A^\ast \mathbf{X}^{\top} \mathbf{Z}_k\mathbf{y}\right)^\top \mathbf{X}^{\top}  \mathbf{Z}_k \mathbf{X} \left( A^\ast \mathbf{X}^{\top} \mathbf{Z}_k \mathbf{y}\right) \\
    &= \mathbf{y}^\top \mathbf{Z}_k \mathbf{X} A^{\top\ast}\mathbf{X}^{\top} \mathbf{Z}_k \mathbf{X} A^\ast \mathbf{X}^{\top} \mathbf{Z}_k \mathbf{y}
\end{align}

Since \( A^\ast \) is symmetric matrix, \( A^{\ast\top} = A^\ast \), we have:
\begin{equation}
    {\mathbf{w}_k^\ast}^\top \mathbf{X}^{\top}  \mathbf{Z}_k \mathbf{X} \mathbf{w}_k^\ast = \mathbf{y}^\top \mathbf{Z}_k \mathbf{X}^{}  A^\ast \mathbf{X}^{\top} \mathbf{Z}_k \mathbf{X} A^\ast (\mathbf{X})^\top \mathbf{Z}_k \mathbf{y}
\end{equation}

\paragraph{Term 4: \(\lambda {\mathbf{w}_k^\ast}^\top \mathbf{w}_k^\ast\) }
\begin{align}
    \lambda {\mathbf{w}_k^\ast}^\top \mathbf{w}_k^\ast & = \lambda \left(A^\ast \mathbf{X}^{\top}  \mathbf{Z}_k \mathbf{y}\right)^\top A^\ast \mathbf{X}^{\top}  \mathbf{Z}_k\mathbf{y}\\
    & = \lambda \mathbf{y}^\top \mathbf{Z}_k\mathbf{X}^{} A^\ast A^\ast \mathbf{X}^{\top} \mathbf{Z}_k \mathbf{y}
\end{align}

\paragraph{Term 3 + Term 4:} 
\begin{align}
     &{\mathbf{w}_k^\ast}^\top \mathbf{X}^{\top}  \mathbf{Z}_k \mathbf{X} \mathbf{w}_k^\ast + \lambda {\mathbf{w}_k^\ast}^\top \mathbf{w}_k^\ast \\
     &= \mathbf{y}^\top \mathbf{Z}_k \mathbf{X} A^\ast \mathbf{X}^{\top} \mathbf{Z}_k \mathbf{X} A^\ast \mathbf{X}^{\top}  \mathbf{Z}_k\mathbf{y}+ \lambda \mathbf{y}^\top \mathbf{Z}_k \mathbf{X}^{} A^\ast A^\ast \mathbf{X}^{\top}  \mathbf{Z}_k \mathbf{y}\\
     &= \mathbf{y}^\top \mathbf{Z}_k \mathbf{X} A^\ast[ \mathbf{X}^{\top}  \mathbf{Z}_k \mathbf{X} + \lambda  \mathbf{I}]A^\ast \mathbf{X}^{\top}  \mathbf{Z}_k \mathbf{y}
 \end{align}                                                        
Substituting the simplified terms back to \eqref{eq:lpc_objective_z_expand}:
\begin{align} 
    &\min_{\mathbf{Z}} \sum_{k=1}^K \left[ \mathbf{y}^\top \mathbf{Z}_k\mathbf{y}- 2 \mathbf{y}^\top \mathbf{Z}_k \mathbf{X} A^\ast \mathbf{X}^{\top} \mathbf{Z}_k\mathbf{y}+ \mathbf{y}^\top \mathbf{Z}_k \mathbf{X}^{}  A^\ast[ \mathbf{X}^{\top}  \mathbf{Z}_k \mathbf{X} + \lambda  \mathbf{I}]A^\ast \mathbf{X}^{\top} \mathbf{Z}_k\mathbf{y}\right]. \\ 
    &\min_{\mathbf{Z}} \sum_{k=1}^K \left[ \mathbf{y}^\top \mathbf{Z}_k\mathbf{y}-  \mathbf{y}^\top \mathbf{Z}_k \mathbf{X} A^\ast [2\mathbf{X}^{\top} \mathbf{Z}_k\mathbf{y}- [ \mathbf{X}^{\top}  \mathbf{Z}_k \mathbf{X} + \lambda  \mathbf{I}]A^\ast \mathbf{X}^{\top}  \mathbf{Z}_k \mathbf{y}] \right]. \\
        &\min_{\mathbf{Z}} \sum_{k=1}^K \left[ \mathbf{y}^\top \mathbf{Z}_k\mathbf{y}-  \mathbf{y}^\top \mathbf{Z}_k \mathbf{X}^{\top}  A^\ast [2\mathbf{I}- (\mathbf{X}^{\top} \mathbf{Z}_k \mathbf{X} + \lambda  \mathbf{I})A^\ast]\mathbf{X}^{\top}  \mathbf{Z}_k\mathbf{y}\right] \label{eq:42}.
\end{align}
Since $ \sum_{k=1}^K \mathbf{Z}_k = I$ and  $\mathbf{y}$ is a constant column vector, \eqref{eq:42} can be simplified into a maximum problem:
\begin{equation}
        \max_{\mathbf{Z}} \sum_{k=1}^K \left[ \mathbf{y}^\top \mathbf{Z}_k \mathbf{X}^{\top}  A^\ast [2\mathbf{I}- (\mathbf{X}^{\top} \mathbf{Z}_k \mathbf{X} + \lambda  \mathbf{I})A^\ast]\mathbf{X}^{\top}  \mathbf{Z}_k\mathbf{y}\right] \label{eq:43}.
\end{equation}

\section{Synthetic Data Setup in Experiments}\label{appendix_ground_truth}
In this appendix, we describe the experimental setup for each research question addressed in the manuscript. 

\textbf{RQ1: Time vs. Error Scalability Analysis}
The standard deviation of noise in the target variable is fixed at \( \sigma = 3.5 \) across all cluster settings to simulate a high-noise dataset with overlapping target ranges. The dimensionality of feature variables is set to 2 to ensure a feasible runtime in this high-noise environment.
\begin{itemize}
    \item \textbf{$K = 2$}: \(\mathbf{w}_1 = [0.2, 0.4, -10]^\top\) \(\mathbf{w}_2 = [-0.2, -0.4, 10]^\top\)
    \item \textbf{$K = 3$:}
    \begin{itemize}
     \item \(\mathbf{w}_1 = [-1.4359, -1.0259, -1.5496, -10]^\top\) 
     \item \(\mathbf{w}_2 = [1.5646, 1.5796, 1.6696, 1]^\top\)
     \item \(\mathbf{w}_3 = [0.1000, 0.1000, 0.1000, 10]^\top\)
    \end{itemize}
    \item  \textbf{$K = 4$:}
        \begin{itemize}
     \item \(\mathbf{w}_1 = [-1.4359, -1.0259, -1.5496, -10]^\top\) 
     \item \(\mathbf{w}_2 = [1.5646, 1.5796, 1.6696, 1]^\top\) 
     \item \(\mathbf{w}_3 = [0.1000, 0.1000, 0.1000, 10]^\top\)
     \item \(\mathbf{w}_4 =  [-3.4360 , -3.02593, -3.5497, 5]^\top\)
    \end{itemize}
\end{itemize} 

\textbf{RQ2: Performance Analysis and Algorithm Properties on Synthetic Datasets}  
All parameters are fixed as follows, except for the parameter designed to change in each experiment: feature dimensionality \( D = 3 \), sample size \( N = 120 \) to ensure feasible runtime for GlobalOpt under 2 hours. For experiments varying feature dimensionality, \( N = 60 \) for \( K = 2 \) and \( N = 42 \) for \( K = 3 \) to ensure feasible in higher dimension. The standard deviation of Gaussian noise in the target variable is set to \( \sigma = 0.5 \). 

Below are the ground truth regression coefficients for varying feature dimensionality experiments, up to \( D = 16 \). Experiments with \( j < 16 \) features use the first \( j \) coefficients for each cluster.

\begin{itemize}
    \item $K = 2$: \begin{itemize}
        \item \(\mathbf{w}_1 = [-1.2668, -1.6211, -1.5291, -1.1345, -1.5135, -1.1844, \\  -1.7853, -1.8539, 
        -1.4942, -1.8465,
        -1.0797, -1.5052, -1.0652, \\ -1.4281, -1.0965, 
        -1.127, 10]^\top\)
        \item \(\mathbf{w}_2 = [1.4032, 1.7739  , 1.8930, 1.7796, 1.6501,\\
        1.5322, 1.7982, 1.3596, 1.5170, 1.49476, \\ 
        1.6131, 1.2063, 1.4200, 1.8377 , 1.2992,
        1.0354, -10]^\top\)
    \end{itemize}

        \item $K = 3$: \begin{itemize}
        \item \(\mathbf{w}_1 = [-1.2668, -1.6211, -1.5291, -1.1345, -1.5135, \\  -1.1844, -1.7853, -1.8539, -10]^\top\)
        \item \(\mathbf{w}_2 = [1.4032, 1.7739  , 1.8930, 1.7796, 1.6501, 1.5322, 1.7982, 1.3596, 1]^\top\)
        \item \(\mathbf{w}_3 = [0.1000,0.1000,0.1000,0.1000,0.1000,0.1000,0.1000,0.1000, 10]^\top\)
    \end{itemize}
\end{itemize}

\newpage
\section{Analysis and Evaluation of Choice of $\alpha$ under Balanced and Imbalanced Cluster Size Setting} \label{imbalanced}

\subsection{Justification of Choice on $\alpha = 1/K$ for Balanced Cluster Setting}

By definition of the non-separable feature space, the clusterwise uncentered covariance,
\begin{equation}
X^{\top} Z_k X,
\end{equation}
computed over the samples assigned to cluster $k$, differs across clusters only in scale, proportional to the number of samples $n_k = \sum_{i} Z_{k,i,i}$ in that cluster. Formally,
\begin{equation}
X^{\top} Z_k X = \sum_{i \in \mathcal{C}_k} x_i x_i^{\top},
\end{equation}

Under the balanced cluster-size assumption used in our non-separable setting (i.e., $n_k \approx N/K$ for all $k$), the uncentered covariance structure is preserved across clusters, implying
\begin{equation}
X^{\top} Z_1 X \;\approx\; X^{\top} Z_2 X \;\approx\; \cdots \;\approx\; X^{\top} Z_K X.
\end{equation}
Summing these clusterwise covariances over all $K$ clusters recovers the global uncentered covariance:
\begin{equation}
X^{\top} X = \sum_{k=1}^{K} X^{\top} Z_k X \;\approx\; K\, X^{\top} Z_k X,
\end{equation}
for any cluster $k$ under balanced partitioning. Rearranging yields the approximation
\begin{equation}
X^{\top} Z_k X \;\approx\; \frac{1}{K} \, X^{\top} X.
\end{equation}

Since the LPC-NS approximation replaces $X^{\top} Z_k X$ with a scaled version of the global covariance, we set
\begin{equation}
\alpha_k = \frac{1}{K},
\end{equation}
leading to the approximate inverse operator used in (7):
\begin{equation}
\bigl( \alpha_k X^{\top} X + \lambda I \bigr)^{-1}
\;\approx\;
\bigl( X^{\top} Z_k X + \lambda I \bigr)^{-1}.
\end{equation}

\subsection{Empirical Evaluation on Imbalanced Cluster Setting with $\alpha$ Tuning }

When cluster sizes are imbalanced, the assumption $\alpha_k = \frac{1}{K} \quad \forall k \in \{1,\ldots,K\}$ used in approximating the cluster-wise uncentered covariance $\alpha_k X^{\top} X$ for non-separable data may no longer hold. In such cases, clusters contribute unequally to the global uncentered covariance $X^{\top} X$: smaller clusters contribute proportionally less, whereas larger clusters contribute more.
To address this, we employ an iterative \emph{$\alpha$-tuning} procedure that selects the cluster imbalance hyperparameter $\alpha_k$ based on the LPC-NS-MIP and LPC-NS-QPBO objectives evaluated under different candidate values. 

\begin{figure}[h!]
    \centering
    \includegraphics[width=0.5\linewidth]{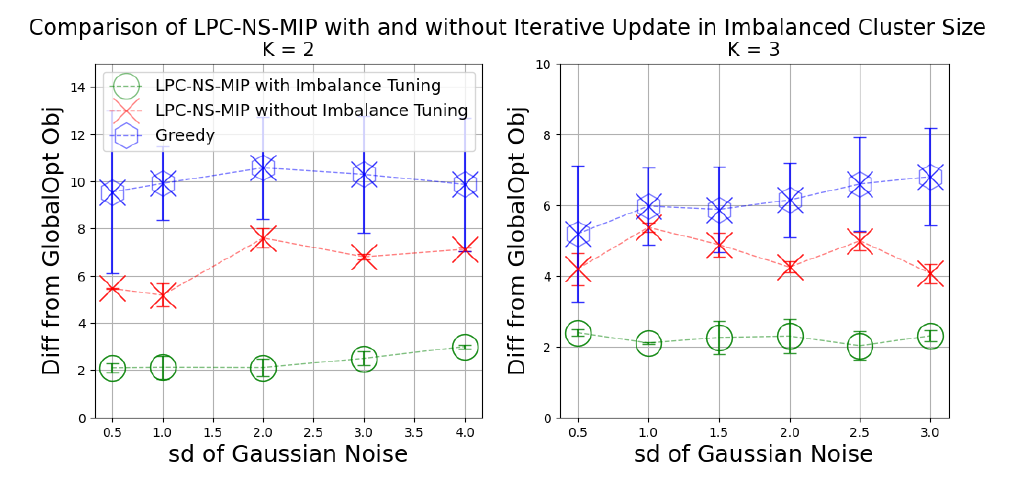}
    \caption{ Difference from the GlobalOpt objective when using LPC-NS-MIP with and without the additional iterative approach to update the cluster imbalance hyperparameter $K$ on a dataset with an imbalanced cluster size.}
    \label{fig:imbalanced}
\end{figure}

We empirically evaluate the effect of $\alpha$-tuning in both LPC-NS-MIP and LPC-NS-QPBO (cf. Fig.~\ref{fig:imbalanced}). The experiments follow the same synthetic data generation process described in Sec.~\ref{sec:experiment_setup}, except that cluster sample sizes are intentionally imbalanced, resulting in differing ground truth uncentered covariance magnitudes across clusters. We observe that the Greedy method exhibits high variability across initializations and consistently yields the largest deviation from the global optimum. In contrast, both LPC-NS-MIP and LPC-NS-QPBO with $\alpha$-tuning achieve performance comparable to, or better than, the balanced setting.

\newpage
\section{Optimization Time Analysis of Dedicated PBO Solvers in Gurobi Against Our Proposed QPBO Model} \label{app:PBO_solver_runtime}

\noindent
We conducted a performance analysis comparing dedicated PBO solvers---specifically Gurobi~\citep{gurobi} and the SCIP Optimization Suite~\cite{BolusaniEtal2024ZR}---against our proposed scalable QBPO method that uses Gurobi's MIP solver, encoding identical problem instances in the OPB format under uniform conditions. The results highlight the runtime limitations of specialized solvers, underscoring the computational advantages of our LPC-NS-QPBO (referred to as LPC-NS-QPBO-MIP-Gurobi in this section to distinguish the implementation approach) method for large-scale instances. 

\begin{figure}[h]
    \centering
    \includegraphics[width=0.6\linewidth]{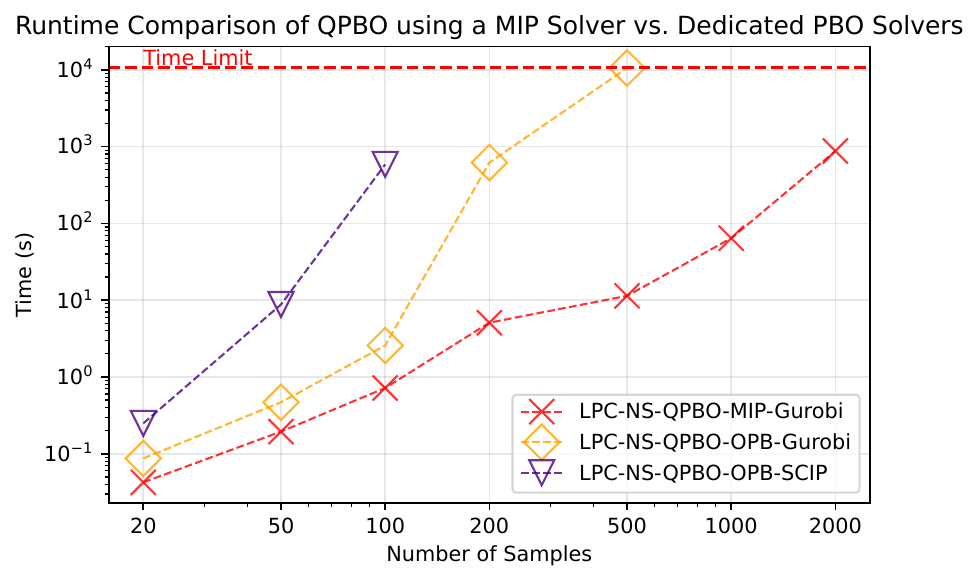}
    \caption{Runtime (y-axis) of different QBPO implementations as the number of samples (x-axis) increases. LPC-NS-QPBO-MIP-Gurobi (red cross) demonstrates significantly better scalability compared to the Gurobi PBO implementation (orange diamond), which reaches the limit of 3 hours at \( N = 500 \), and SCIP PBO (purple triangle), which reaches the limit at \( N = 100.\) }
    \label{fig:pbo_comparison}
\end{figure}

As illustrated in Fig.~\ref{fig:pbo_comparison}, we evaluated different implementations of the same QPBO formulation for \(n_{\mathrm{values}} \in \{20, 70, 100, 200, 500, 1000, 2000\}\), dimensionality \(D=1\), \(K=2\), and noise \( \sigma = 1.5\). Among the PBO solvers tested, the SCIP PBO solver exhibited the least scalability, followed by Gurobi, as both these methods reached the runtime limit at \(N = 100\) and \(N = 500\), respectively, whereas our method demonstrated significant scalability improvements in the number of instances.

\newpage
\section{Additional Synthetic Experiments} \label{app:outliers}

\subsection{Performance Analysis of Varying Proportion of Outliers}

\begin{figure}[!h]
    \centering
    \includegraphics[width=0.8\linewidth]{plots/RQ2_Outlier_k=2.pdf}
    \includegraphics[width=0.8\linewidth]{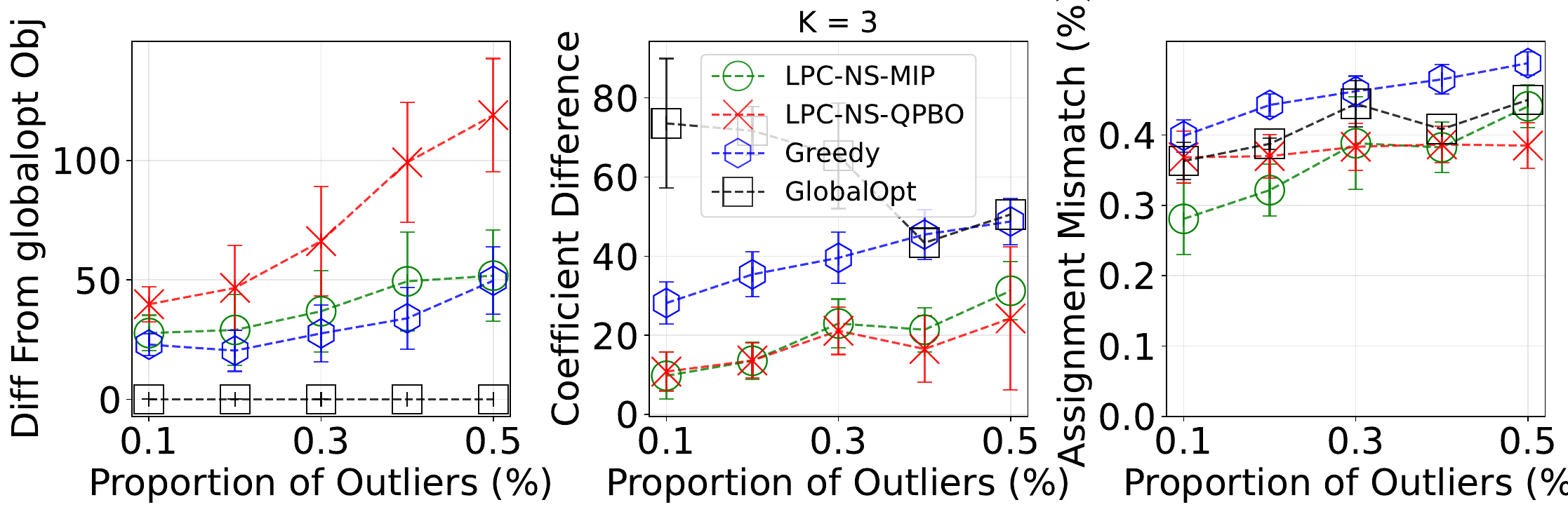}
\caption{\textbf{RQ2 Proportion of Outliers} The difference from the globally optimal objective, ground truth regression coefficients, and ground truth cluster assignments across different methods as the proportion of outliers in target variables increases. The 95\% confidence interval is visualized as straight lines in the plots. }
    \label{fig:outlier_3}
\end{figure}

Fig.~\ref{fig:outlier_3} report the performance of different methods as the proportion of outliers increases for \( K = 2 \) and \( K = 3 \). LPC-NS-MIP maintains similar performance in this setting. 

In \( K = 3 \),  analyzing the recovery of \( \mathbf{w}^{\text{gt}} \) and \( \mathbf{Z}^{\text{gt}} \) (second and third plots) reveals that all methods struggle to capture the true cluster assignments. A possible explanation is that outliers in cluster \( k \) often overlap with the target variable range of other clusters, leading all methods to misclassify these outliers to minimize SSE. Consequently, none of the methods can reliably recover the correct cluster assignments in high-outlier scenarios.

\subsection{Performance Analysis for $K = 5$}\label{app:high_cluster_experiment}

In this section, we clarify the rationale for focusing our main experimental analysis on $K = 2$ and $K = 3$. First, for higher cluster counts ($K > 3$), the \textsc{GlobalOpt} formulation becomes computationally infeasible to solve within reasonable time limits, as both the number of binary indicator variables and the number of regression coefficients scale combinatorially with $K$. This exponential growth severely restricts the applicability of global MIP solvers in higher-cluster settings.

Second, to empirically evaluate the behavior of different methods beyond this range, we conduct an additional experiment on synthetic data with $K = 5$. Since \textsc{GlobalOpt} is infeasible in this regime, we compare methods directly based on their final clustering objective values rather than deviation from the global optimum.

\begin{figure}[!h]
    \centering
    \includegraphics[width=\linewidth]{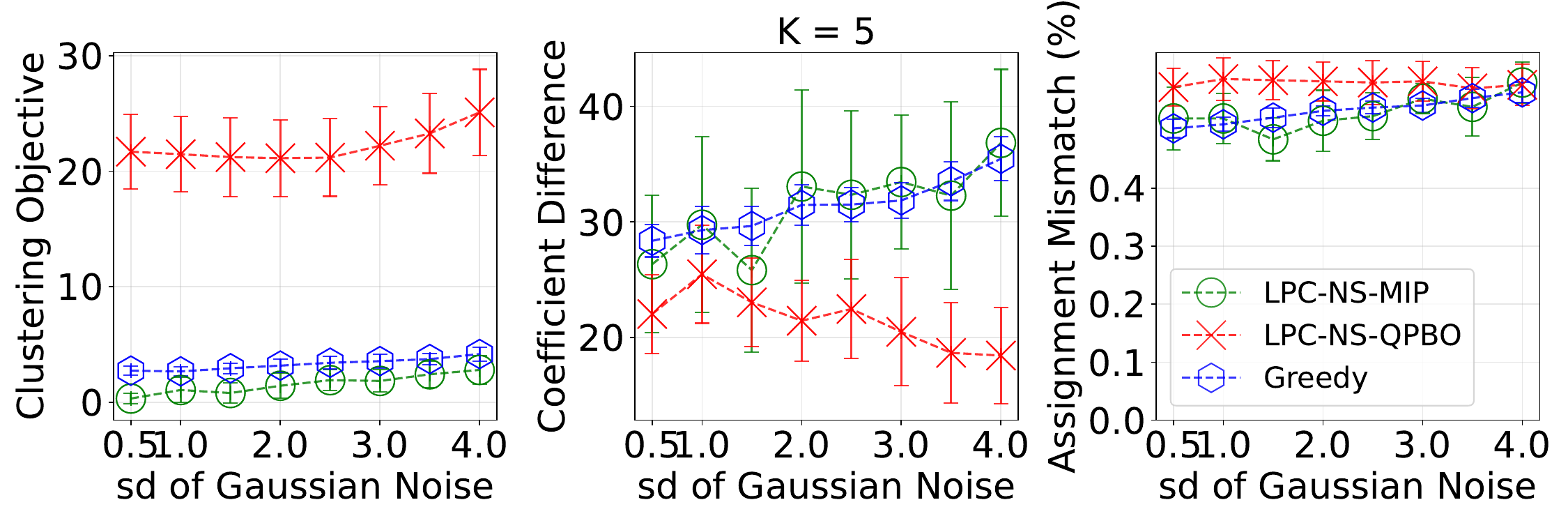}
    \caption{\textit{RQ2 Proportion of Outliers.} The deviation from the globally optimal objective, ground truth regression coefficients, and ground truth cluster assignments for different methods as the proportion of outliers increases. The shaded regions indicate the $95\%$ confidence interval.}
    \label{fig:K = 5}
\end{figure}

Figure~\ref{fig:K = 5} reports the results for $K = 5$ using the synthetic data generation procedure described in Section~\ref{sec:experiment_setup}. In this setup, the five clusters follow distinct linear relationships: two exhibit positive slopes, two negative slopes, and one follows an uncorrelated (flat) pattern.

We observe that, in higher-cluster scenarios, the LPC objective may admit multiple equivalent cluster assignments for a single sample. In particular, different clusters can achieve identical SSE contributions for a given sample, resulting in ambiguous assignments. This ambiguity increases the cluster assignment mismatch and leads to large discrepancies in recovered regression coefficients relative to the ground truth for all methods. Consequently, as $K$ increases, the interpretability and analytical usefulness of LPC evaluations diminish.

Moreover, in practice, encountering more than five non-separable and mutually overlapping clusters is highly uncommon (cf. App.~G, where we assess separability using ground truth categorical labels across real datasets). In such cases, simpler clustering procedures (e.g., CLR or any feature-based clustering approach) can first partition separable subsets of the data. LPC is then best applied to the residual, genuinely non-separable substructure, aligning with its intended use case.

Nevertheless, even though the direct objective comparison in this experiment is not informative due to misalignment with the true cluster structure, \textsc{LPC-NS-MIP} consistently achieves lower clustering objective values than \textsc{Greedy}. In contrast, \textsc{LPC-NS-QPBO} performs poorly in this setting, reflecting the limitations of its more aggressive approximation when $K$ becomes large.

\newpage
\section{Descriptions and Precoessing Steps of Real Datasets}
\label{sec:app_real_datasets}
\subsection{Detail Descriptions of Real Datasets} 
This appendix provides a detailed description of each real-world dataset used in \textbf{RQ3} and the preprocessing steps applied before experimentation. The `Dataset Name' and `UCI ID' columns in Tab.~\ref{tab:real_data_description} list the datasets from the UCI Machine Learning Repository along with their unique IDs for retrieving raw data. We selected datasets accessible via the UCI Python API that contained at least one categorical feature for later processing. For datasets with more than 600 samples, we randomly sampled 600 instances to ensure LPC-NS-MIP could be optimized within a feasible runtime.  

Standard preprocessing steps were applied, including one-hot encoding for categorical variables and z-score normalization. For datasets with more than 15 features, we used F-statistics to select the top 15 features with a p-value $ \leq 0.05 $. The `Features' column in Tab.~\ref{tab:real_data_description} reports the number of feature variables retained for each dataset.  

Each dataset was partitioned into $ K $ groups based on values in the categorical feature. For each cluster $ k $, a ridge regression model was fitted separately to obtain clusterwise coefficients $ \mathbf{w}_k $. We then computed the pairwise regression coefficient differences using the $ \ell_2 $ norm to identify the two clusters with the most distinct regression relationships:  

\begin{equation}
    \max_{i,j}\|\mathbf{w}_i - \mathbf{w}_j\|_2 \quad \forall i, j \in \{1,\dots,k\}
\end{equation}\label{pairwise_weight_difference}

This approach ensures that the resulting dataset contains at least two clusters with distinct linear relationships, aligning with the LPC problem setting. Additionally, it provides reference cluster assignments (i.e., the categorical feature used for partitioning) and regression coefficients, enabling evaluation of each method’s ability to recover the true underlying relationships. However, this assignment and regression coefficients are not ground truth since they are assigned manually. But, it provides a reference comparison between each method under a controllable setting.

For datasets with multiple categorical features, we repeated this process for each categorical feature and selected the one yielding the largest $ \ell_2 $ norm in  \eqref{pairwise_weight_difference}. The categorical feature used for splitting is reported in the `Categorical Feature' column of Tab.~\ref{tab:real_data_description}.  

After selecting the categorical feature and the top two clusters with the largest regression coefficient differences, all other samples and the categorical feature used for partitioning were removed. The final number of samples in each dataset is reported in the `Samples' column of Tab.~\ref{tab:real_data_description}, while the sample sizes for each cluster are listed in the `Cluster 1 Size' and `Cluster 2 Size' columns.

\begin{table}[!h]
    \centering
    \scalebox{0.8}{
\begin{tabular}{ll|rr|lrr}
\toprule
Dataset Name & UCI ID & Samples & Features & Categorical Feature & Cluster 1 Size & Cluster 2 Size \\
\midrule
Stock Portfolio Performance  & 390 & 126 & 6 & period & 63 & 63 \\
Servo  & 87 & 69 & 6 & motor & 33 & 36 \\
Solar Flare  & 89 & 212 & 14 & largest spot size & 125 & 87 \\
Productivity Prediction  & 597 & 204 & 12 & day & 102 & 102 \\
Heart Failure Prediction  & 519 & 299 & 11 & sex & 194 & 105 \\
Liver Disorders  & 60 & 345 & 5 & selector & 145 & 200 \\
Student Performance  & 320 & 56 & 15 & Fjob & 33 & 23 \\
Parkinsons Telemonitoring  & 189 & 600 & 15 & sex & 396 & 204 \\
Facebook Metrics  & 368 & 433 & 13 & Type & 426 & 7 \\
Infrared Thermography  & 925 & 23 & 15 & Age & 15 & 8 \\
\bottomrule
\end{tabular}
}
\caption{Description of real-world datasets used in \textbf{RQ3}. The `Dataset Name' and `UCI ID' columns provide the dataset names and corresponding unique identifiers in the UCI Machine Learning Repository. `Features' indicates the total number of feature variables after preprocessing. `Categorical Feature' specifies the feature used to define initial clusters. `Samples' reports the total number of selected samples, while `Cluster 1 Size' and `Cluster 2 Size' indicate the number of samples in each selected cluster after filtering.}
    \label{tab:real_data_description}
\end{table}
\subsection{Analysis of Separability between Ground Truth Clusters}
In practice, cases where three or more truly distinct regression relationships exist within the same dataset are relatively rare.  Tab.~\ref{tab:real_data_regression_difference} shows the cluster-wise $\ell_2$ distance of the regression coefficients between all possible pairs of clusters based on the selected categorical feature. These results reinforce the claim that meaningful distinctions—i.e., large differences in regression weights—typically only occur for a single pair of feature values in each dataset.

 \begin{table}[!h]
    \centering
    \scalebox{0.8}{
\begin{tabular}{l|l}
\toprule
Dataset Name & List of $\ell_2$ Distance Between All Pairs of Cluster's Regression Coefficients\\
\midrule
Stock Portfolio Performance  & 0.189, 0.544, 0.548, 0.818, 0.266, \textbf{4.341}\\
Servo  & 0.013, 0.122, 0.163, 0.165, 0.171, 0.2, 0.21, 0.243, 0.282, \textbf{0.286} \\
Solar Flare & 0.064, 0.106, 0.135, \textbf{5820653786634.189} \\
Productivity Prediction  & 0.197, 0.266, 0.342, 0.832, 0.853, 1.026,1.94, 2.132, 2.138,
2.958, \textbf{3784167278436.165} \\
Heart Failure Prediction  & 0.063 \\
Liver Disorders  & 0.087 \\
Parkinsons Telemonitoring  & 35.496 \\
Facebook Metrics  & $1.2364 \times 10^{-15}$ \\
Infrared Thermography  & 0.292, 0.764, 0.807, 0.948, 0.972, 1.493, 2.259, 2.405, 1.469, 0.263, 0.439, \textbf{4.409} \\
\bottomrule
\end{tabular}
}
\caption{The pairwise $\ell_2$ distance of the regression coefficients between all possible pairs of clusters, based on the value in the selected categorical features for cluster assignment. \textsc{Student Performance (320)} and \textsc{Total Rental Bike (275)} are not included in the table due to the total possible pairs exceeding the size limit.}
    \label{tab:real_data_regression_difference}
\end{table}

\newpage
\subsection{Regression Coefficients Difference and Clustering Assignment Mismatch in the Real Datasets}
\label{sec:coef_diff_assign_mismatch}

These regression weight differences in Tab.~\ref{tab:weight_real} and cluster assignment mismatch in Tab.~\ref{tab:cluster_real} from 10 real datasets stem from how the data was curated and may not necessarily reflect the optimal cluster labels. However, the lower objective error achieved by our methods suggests that alternative methods were unlikely to produce better labels.

\begin{table}[!h]
    \centering
\scalebox{0.8}{
\begin{tabular}{|c|rrr|}
\toprule
\multicolumn{1}{|c|}{\textbf{Dataset}} & \multicolumn{3}{c|}{\textbf{Regession Coefficient Difference}} \\
Dataset (UCI Repository ID) & LPC-NS-QPBO & LPC-NS-MIP & Greedy \\
\midrule
Stock Portfolio Performance (390) & \textbf{18.3930} & 57.3053 & $49.7234 \pm 0.2030$ \\
Servo (87) & 5.0118 & \textbf{4.7162} & $5.1693 \pm 0.2684$ \\
Solar Flare (89) & \textbf{3.4549} & 5.3680 & $3.8343 \pm 0.1907$ \\
Productivity Prediction (597) & 25.1202 & \textbf{9.7577} & $22.2822 \pm 1.8296$ \\
Heart Failure Prediction (519) & 2.5921 & \textbf{1.8778} & $2.9453 \pm 0.3060$ \\
Liver Disorders (60) & 1.9037 & 1.7529 & $\mathbf{1.5192 \pm 0.0702}$ \\
Student Performance (320) & 7.2744 & 8.1516 & $\mathbf{5.5334 \pm 0.2439}$ \\
Parkinsons Telemonitoring (189) & 1268.0732 & 1450.8159 & $\mathbf{76.4961 \pm 0.42930}$ \\
Facebook Metrics (368) & \textbf{0.0000} & 0.6830 & $0.5997 \pm 0.0133$ \\
Infrared Thermography (925) & 221.3324 & 244.7812 & $\mathbf{193.3577 \pm 0.7206}$ \\
\bottomrule
\end{tabular}
}
   \caption{Regression coefficient differences across three methods. The differences are computed using regression coefficients obtained by fitting a regression model based on cluster assignments derived from the categorical feature used for partitioning. For Greedy, the 95\% confidence interval is reported using 50 different initialization values.}
    \label{tab:weight_real}
\end{table}
\begin{table}[!h]
    \centering
    \scalebox{0.8}{
    \begin{tabular}{|c|rrr|}
    \toprule
\multicolumn{1}{|c|}{\textbf{Dataset}} & \multicolumn{3}{c|}{\textbf{Cluster Assignment Mismatch (\%)}} \\
Dataset (UCI Repository ID) & LPC-NS-QPBO & LPC-NS-MIP & Greedy \\
\midrule
Stock Portfolio Performance (390) & 0.277800 & \textbf{0.087300} & $0.1983 \pm0.0346$ \\
Servo (87) & 0.4493 & \textbf{0.3768} & $0.402 \pm0.0119$ \\
Solar Flare (89) & 0.4198 & \textbf{0.2217} & $0.3188 \pm0.0343$ \\
Productivity Prediction (597) & 0.4706 & 0.4902 & $\mathbf{0.4655 \pm0.0059}$ \\
Heart Failure Prediction (519) & \textbf{0.4448} & 0.4582 & $0.4595 \pm0.006$ \\
Liver Disorders (60) & \textbf{0.4493} & 0.4841 & $0.4744 \pm0.006$ \\
Student Performance (320) & 0.4643 & 0.48210& $\mathbf{0.4307 \pm0.0149}$ \\
Parkinsons Telemonitoring (189) & 0.4800 & \textbf{0.4583} & $0.4783 \pm0.0049$ \\
Facebook Metrics (368) & \textbf{0.0162} & 0.3972 & $0.0437 \pm0.002$ \\
Infrared Thermography (925) & 0.3913 & \textbf{0.3043} & $0.4243 \pm0.0141$ \\
\bottomrule
\end{tabular}
}
    \caption{This table presents the clustering assignment mismatches across three methods. The mismatch values measure the deviation of cluster assignments obtained by each method from the reference assignments derived from the categorical feature used for partitioning. The 95\% confidence interval using 50 initialization value of Greedy is reported.}
    \label{tab:cluster_real}
\end{table}
% \newpage

\end{document}